\documentclass{article}

\PassOptionsToPackage{numbers}{natbib}

     \usepackage[preprint]{neurips_2020}

\usepackage[utf8]{inputenc} 
\usepackage[T1]{fontenc}    
\usepackage{hyperref}       
\usepackage{url}            
\usepackage{booktabs}       
\usepackage{amsfonts}       
\usepackage{nicefrac}       
\usepackage{microtype}      
\usepackage{amsmath}
\usepackage{graphicx}
\usepackage{subfig}
\usepackage{amssymb}

\title{RIC-CNN: Rotation-Invariant Coordinate Convolutional Neural Network}

\author{
  Hanlin Mo~~~~~~Guoying Zhao\thanks{Corresponding Author}\\
  Center for Machine Vision and Signal Analysis (CMVS), University of Oulu\\
  \texttt{\{hanlin.mo, guoying.zhao\}@oulu.fi} \\
}

\begin{document}

\maketitle

\begin{abstract}
In recent years, convolutional neural network has shown good performance in many image processing and computer vision tasks. However, a standard CNN model is not invariant to image rotations. In fact, even slight rotation of an input image will seriously degrade its performance. This shortcoming precludes the use of CNN in some practical scenarios. Thus, in this paper, we focus on designing convolutional layer with good rotation invariance. Specifically, based on a simple rotation-invariant coordinate system, we propose a new convolutional operation, called Rotation-Invariant Coordinate Convolution (RIC-C). Without additional trainable parameters and data augmentation, RIC-C is naturally invariant to arbitrary rotations around the input center. Furthermore, we find the connection between RIC-C and deformable convolution, and propose a simple but efficient approach to implement RIC-C using Pytorch. By replacing all standard convolutional layers in a CNN with the corresponding RIC-C, a RIC-CNN can be derived. Using MNIST dataset, we first evaluate the rotation invariance of RIC-CNN and compare its performance with most of existing rotation-invariant CNN models. It can be observed that RIC-CNN achieves the state-of-the-art classification on the rotated test dataset of MNIST. Then, we deploy RIC-C to VGG, ResNet and DenseNet, and conduct the classification experiments on two real image datasets. Also, a shallow CNN and the corresponding RIC-CNN are trained to extract image patch descriptors, and we compare their performance in patch verification. These experimental results again show that RIC-C can be easily used as drop in replacement for standard convolutions, and greatly enhances the rotation invariance of CNN models designed for different applications.  
\end{abstract}

\section{Introduction}
\label{section:1}
\begin{figure}
	\centering
	\subfloat[Handwritten digit recognition.]
	{\includegraphics[height=14mm,width=35mm]{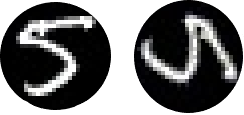}\label{figure:1(a)}\hfill}~~~~~
	\subfloat[Airplane type recognition in aerial images.]
	{\includegraphics[height=14mm,width=35mm]{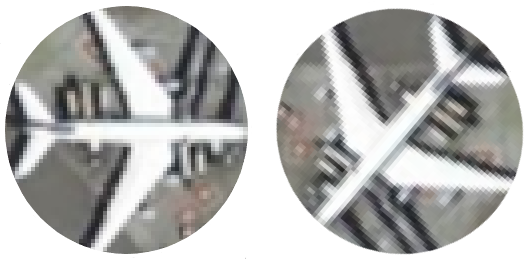}\label{figure:1(b)}\hfill}~~~~~
	\subfloat[Image patch matching.]
	{\includegraphics[height=14mm,width=35mm]{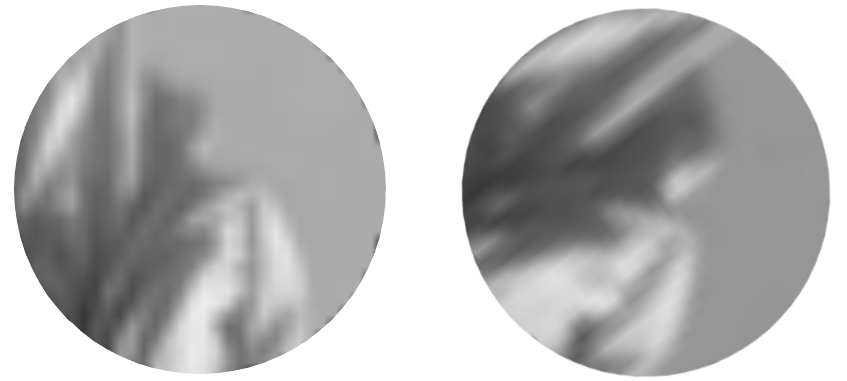}\label{figure:1(c)}\hfill}~~~~~
	\caption{In many practice tasks, the rotation invariance of image features is very important.}\label{figure:1}
\end{figure}

How to extract effective features of images is one of the most fundamental problems in the fields of computer vision and pattern recognition. A desirable feature should have the ability to capture intrinsic information of an object of interest in a given image. In most cases, we can observe and take pictures of an object from different viewpoints. This implies that the feature should at least be invariant to various types of spatial deformations caused by imaging geometry, such as two-dimentional translation, rotation, similarity transform, affine transform and so on. Among these image transformation models, two-dimensional rotation plays a very important role. This widely used transform contains only one parameter, namely the angle of rotation, and it is a key part of more complicated transforms. The rotation invariance of image features is essential for many practice tasks, such as handwritten digit recognition, aircraft type recognition in remote sensing images and image patch matching (see Figure \ref{figure:1}). As a result, for decades, constructing invariant features to image rotation has attracted considerable attention. Based on different mathematical tools, researchers designed numerous rotation-invariant hand-crafted features for image retrieval, object recognition, textural image classification and image patch matching, including Scale Invariant Feature Transform (SIFT)\cite{1}, Local Binary Patterns (LBP)\cite{2}, Rotation-Invariant Feature Transform (RIFT)\cite{3}, Local Intensity Order Patterns (LIOP)\cite{4} and so on. 

Since 2012, deep-learning architectures, especially Convolutional Neural Network (CNN), have been successfully applied in many computer vision tasks and significantly outperform most of image hand-crafted features. However, traditional CNN models are not invariant to image rotation, and even small rotations of an input image can drastically degrade their performance. In fact, previous studies have proved that classical convolutional operation just has the invariance to image translation, which is achieved by local connectivity and spatial parameter sharing. A simple way to cope with image rotation is to train CNNs by using data augmentation. Unfortunately, this approach not only increases the cost of training time but also wastes computation to learn a lot of redundant weights. For example, the visualisation of first-layer weights in a CNN reveals that many of them are rotated copies of one another \cite{5}. In addition, it is not easy to determine how many rotation versions should be generated for each of training images. Thus, recent research has begun to focuse on designing new network architectures to encode rotation invariance into classical CNNs and has proposed multiple modified models, such as Spatial Transformer Networks (STN) \cite{6}, Transformation-Invariant Pooling (TI-Pooling)\cite{7}, Rotation Equivariant Vector Field Networks (RotEqNet)\cite{8}, Harmonic Networks (H-Net)\cite{9}, Group Equivariant Convolutional Networks (G-CNN)\cite{10} and so on. However, there are three major limitations of these existing CNN models for achieving rotation invariance. First, most of them are just invariant to special rotation angles rather than to arbitrary image rotations. Secondly, some methods require extra trainable parameters to handle image rotation. In addition, due to depending on complex mathmatical concepts and theories, such as group theory, several methods are not intuitive. Thirdly, most of researchers still trained their rotation-invariant CNN models with image data augmentation. It is unreasonable. Based on the experimental results in these papers, we can not determine whether certain modified architecture or just rotation augmentation enhances the invariance of a CNN model to image rotation. More importantly, when using data augmentation, compared with conventional CNNs, rotation-invariant CNNs just obtain insignificant performance increase \cite{11}. 

The goal of our paper is to address these limitations, and the main contributions can be summarized as follows: \textbf{(i)} Using a simple rotation-invariant coordinate system, we design a novel convolutional operation, Rotation-Invariant Coordinate Convolution (RIC-C). By replacing all standard convolutional operations in a CNN with the corresponding RIC-C, we can derive a Rotation-Invariant Coordinate Convolutional Neural Network (RIC-CNN), which is naturally invariant to all rotation angles around the input center. Also, no extra learnable parameters are used to achieve its rotation invariance, which means the number of parameters in RIC-CNN is exactly the same as the original CNN. \textbf{(ii)} We find the connection between RIC-C and deformable convolution \cite{12}, and propose a efficient approach to implement RIC-C using Pytorch. \textbf{(iii)} Based on MNIST dataset \cite{13}, we train RIC-CNN without data augmentation and evaluate its rotation invariance. Most of rotation-invariant CNNs proposed in previous literature are chosen for comparison. The results clearly show that RIC-CNN significantly outperforms the state of the art approaches. Then, RIC-C is successfully deployed to some popular CNN models, including VGG\cite{14}, ResNet\cite{15} and DenseNet\cite{16}, and real image classification is conducted on NWPU VHR-10 \cite{17} and MTARSI datasets\cite{18}. Finally, using triplet loss \cite{19}, a shallow CNN and the corresponding RIC-CNN are trained on UBC benchmark dataset \cite{20} to extract patch descriptors. We compare their performance in patch verification task. The rest of our paper is organized as follows. In Section \ref{section:2}, we formally define rotation-invariant image features and explain why conventional convolution is not invariant to two-dimensional rotation. Section \ref{section:3} summarizes the existing approaches for achieving rotation invariance of hand-crafted image features and CNN models. Section \ref{section:4} is the main contribution of this paper. We define RIC-CNN in detail, and introduce how to implement this novel CNN architecture efficiently. In Section \ref{section:5}, we conduct comprehensive experiments to evaluate the performance of RIC-CNN in different tasks. Finally, Section \ref{section:6} presents our conclusions and future plans.     

\section{Basic Concepts and Definitions}
\label{section:2}
In this section, we will introduce some basic concepts and definitions used in the following sections.

\subsection{Rotation-Invariant Image Features}
\label{section:2.1}
An image can be defined as a multi-channel function $F(X):\Omega\subset\mathbb{R}^{2}\rightarrow\mathbb{R}^{m}$, where $X=\left(x_{1},x_{2}\right)^{T}$ and $F(X)=\left(f_{1}(X),f_{2}(X),...,f_{m}(X)\right)^{T}$. Note that 
\begin{itemize}
\item in this paper, the Cartesian coordinate system of $X$ is established at the center of the image $F(X)$, which is denoted as $X_{c}=\left(x^{c}_{1},x^{c}_{2}\right)^{T}$. This means that we always have $X_{c}=\left(0,0\right)^{T}$.    
\item the positive integer $m$ represents the channel dimension of $F(X)$, and for any $k\in\{1,2,...,m\}$, the scalar function $f_{k}(X)$ acts as mapping from the domain $\Omega$ to $\mathbb{R}$. 
\end{itemize} 
We can instantiate $F(X)$ by setting specific $m$. For example, when setting $m=3$, it can be used to represent arbitrary RGB images.

Further, an image feature can be represented as a function $\Phi$ which maps $F(X)$ to a feature space $\Phi(F(X))\subset\mathbb{R}^{d}$, where $d$ is a positive integer. Basically there are two types of image features, global features and local features. A global feature describes the image $F(X)$ as a whole. More specifically, a $d$-dimensional feature vector $\Phi(F(X))$ is used to generalize the entire object. On the contrary, a local feature is calculated from the local region around a given point $X_{0}=\left(x^{0}_{1},x^{0}_{2}\right)^{T}\in\Omega$, and can be expressed as $\Phi(X_{0},F(X))$. It is suitable to measure local structures. If we calculate the local feature $\Phi(X, F(X))$ for each $X\in\Omega$, we actually obtain a new $d$-channel image. Previous researchers usually called $\Phi(X,F(X))$ the feature map of $F(X)$.   

Let the image $F(X)$ be rotated $\theta$ degrees around the origin $X_{c}$ to $G(Y):\Omega^{'}\subset\mathbb{R}^{2}\rightarrow\mathbb{R}^{m}$, we have 
\begin{equation}\label{equ:1}
G(Y)=F(X),~~\mbox{where}~~
Y=\left(
\begin{array}{c}
	y_{1}\\
	y_{2}\\
\end{array}
\right)
=R_{\theta}X
=
\left(
\begin{array}{cc}
	cos(\theta) & -sin(\theta)\\
	sin(\theta) & cos(\theta)\\
\end{array}
\right)
\left(
\begin{array}{c}
	x_{1}\\
	x_{2}\\
\end{array}
\right)
\end{equation}
Note that the rotation angle $\theta\in[0,2\pi)$ and is measured anticlockwise. The point $Y_{c}=\left(y^{c}_{1},y^{c}_{2}\right)$ denotes the center of $G(Y)$. It is obvious that $Y_{c}=X_{c}$. In Figure \ref{figure:1}, we have shown some pairs of images related by (\ref{equ:1}). Suppose that an image feature $\Phi$ is invariant to image rotation, we have $\Phi(G(Y))=\Phi(F(X))$. Particularly, if $\Phi$ is a local feature, we have $\Phi(Y, G(Y))=\Phi(X, F(X))$, where $Y=R_{\theta}\cdot X$. This indicates that the feature map $\Phi(Y,G(Y))$ is also a rotated version of the original map $\Phi(X,F(X))$. This property is called rotation equivariance of the local feature $\Phi$. Clearly, for a local feature, its rotation invariance naturally ensures its rotation equivariance. Hence, our paper just focuses on achieving rotation invariance of image local features and does not dicuss their rotation equivariance separately.    

\subsection{Conventional Convolution}
\label{section:2.2}
\begin{figure}
	\centering
	\subfloat[The order of $\left(4n^{2}+4n+1\right)$ sample points $P$ on the regular grid $\mathcal{S}$.]
	{\includegraphics[height=50mm,width=52mm]{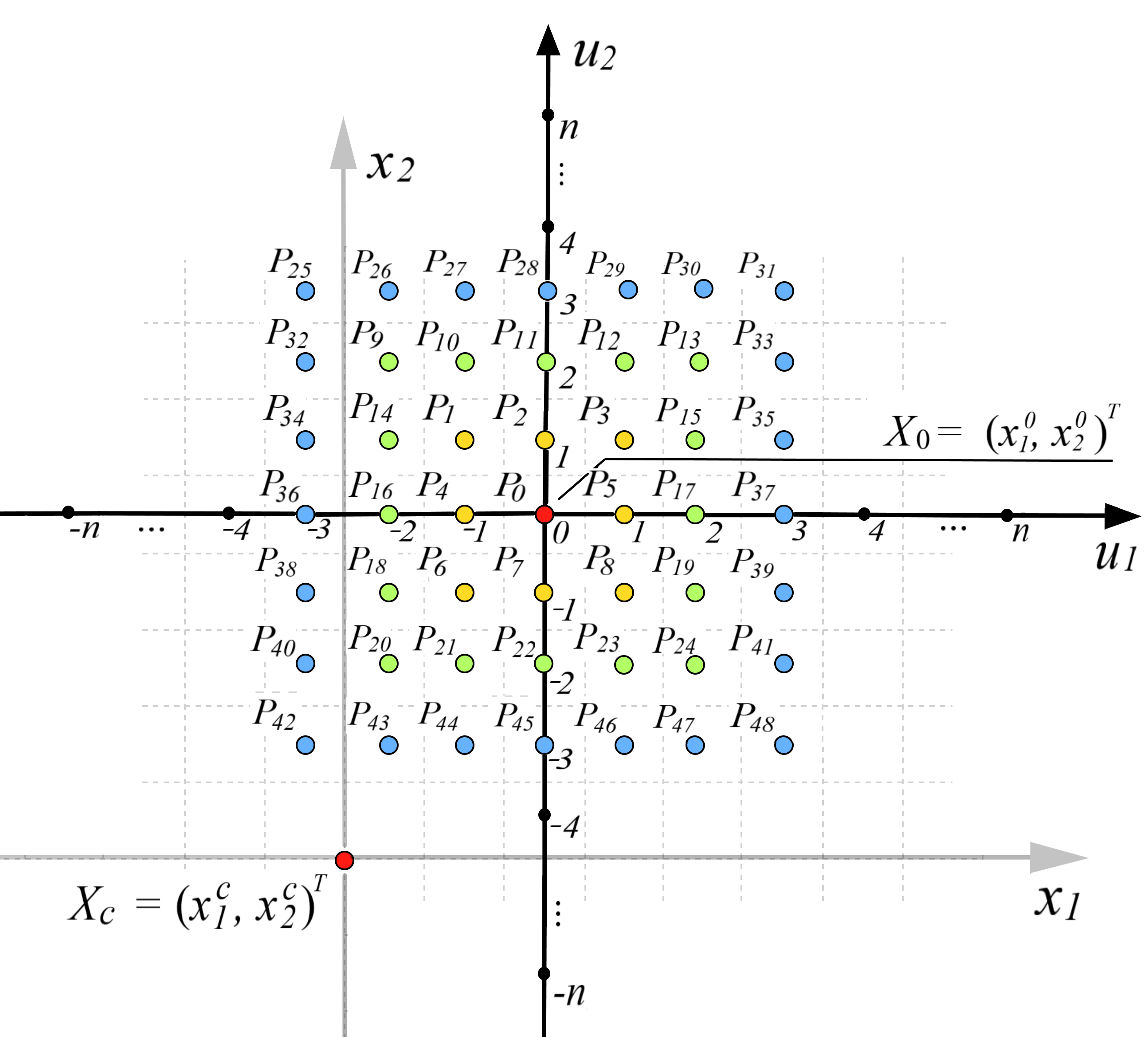}\label{figure:2(a)}\hfill}~~~~~~~~~~~
	\subfloat[The order of $\left(4n^{2}+4n+1\right)$ sample points $Q$ on the rotation-invariant coordinate system $\mathcal{T}$.]
	{\includegraphics[height=50mm,width=52mm]{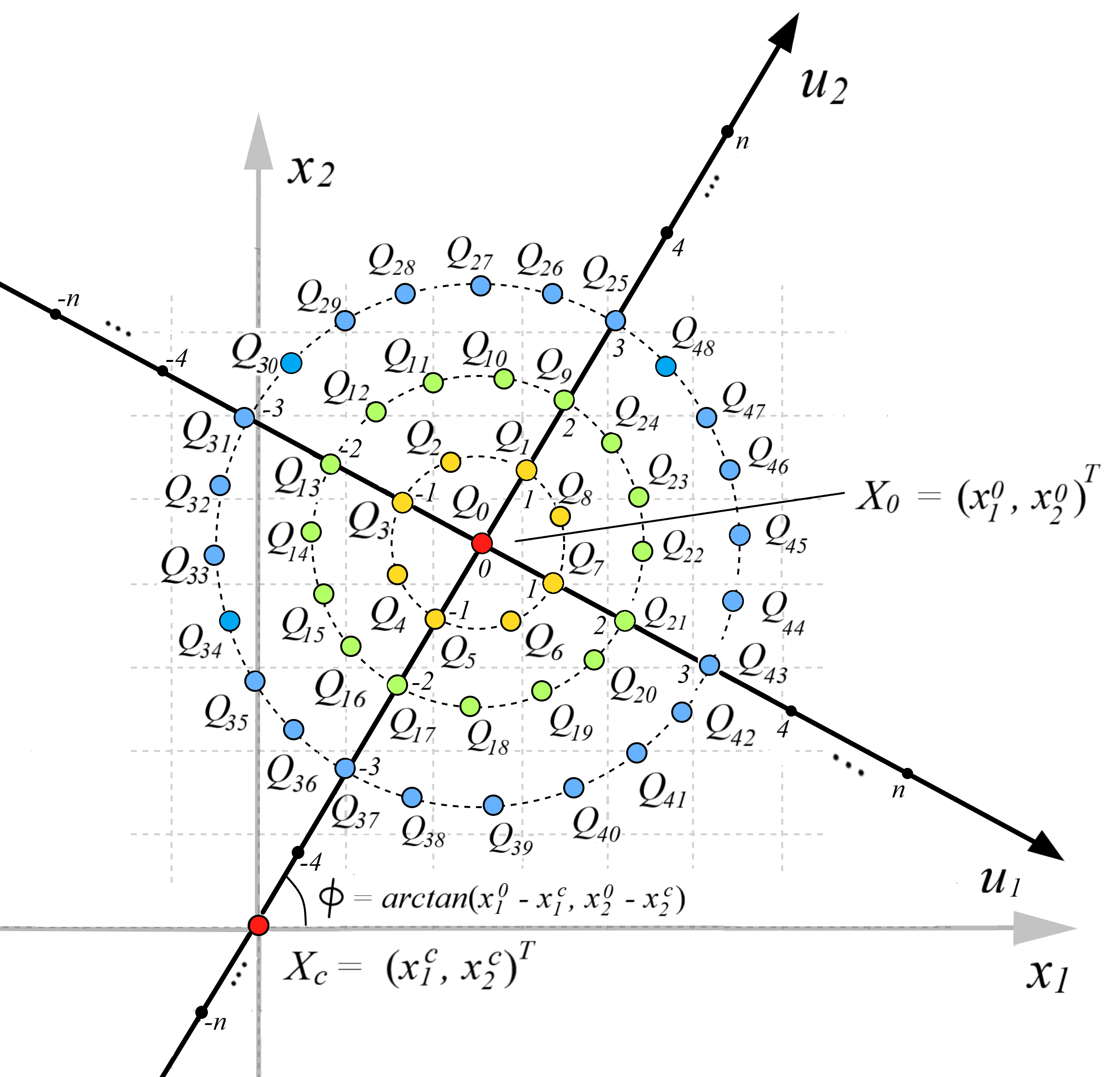}\label{figure:2(b)}\hfill}
	\caption{Two sets of sampling points used for calculating $\Phi_{C}$ and $\Phi_{RIC-C}$, respectively. It is clear that $Q_{\alpha}\in \mathcal{T}$ can be regarded as the corresponding point of $P_{\alpha}\in\mathcal{S}$, where the subscript $\alpha=0,1,2,...,\left(4n^{2}+4n\right)$.}\label{figure:2}
\end{figure}

Convolutional operation is a kind of local feature. For an input $F(X)$ defined above, a convolutional operation $\Phi_{C}$ acting on a given point $X_{0}=\left(x^{0}_{1},x^{0}_{2}\right)^{T}\in \Omega$ can be expressed as
\begin{equation}\label{equ:2}
\Phi_{C}(X_{0},F(X))=\sum_{P\in\mathcal{S}}W(P)\cdot F(X_{0}+P)
\end{equation}
where $W$ represents a learnable convolutional kernel, $P$ enumerates all sample locations on a regular grid $\mathcal{S}$. Specifically, given a non-negative integer $n$, we set the size of the kernel $W$ equal to $(2n+1)\times(2n+1)$, and the regular grid $\mathcal{S}=\{-n,-n+1,\cdots, n-1, n\}\times\{-n,-n+1,\cdots, n-1, n\}$. Only odd-sized convolutional kernels are considered in our paper because the shift problem occurs in even-sized kernels \cite{21}. Note that $\mathcal{S}$ contains $(2n+1)\cdot(2n+1)=\left(4n^{2}+4n+1\right)$ sample points. For example, when setting $n=1$, the kernel size is $3\times 3$ and $\mathcal{S}=\{-1,0,1\}\times\{-1,0,1\}=\{(-1,-1)^{T},(-1,0)^{T}, \cdots, (0,1)^{T},(1,1)^{T}\}$. There are $3\times 3=9$ sample points. The regular grid $\mathcal{S}$ is totally determined by the value of $n$, and it is independent from the position $X_{0}$ where the convolutional operation $\Phi_{C}$ acting. In addition, we can order all sample points as shown in Figure \ref{figure:2(a)}, where the subscript of $P_{\alpha}$ indicates its order, and $\alpha = 0,1,2,...,\left(4n^{2}+4n\right)$.
          
\begin{figure}
	\centering
	\subfloat[Illustration of the sensitivity of $3\times3$ conventional convolution to image rotation.]
	{\includegraphics[height=60mm,width=115mm]{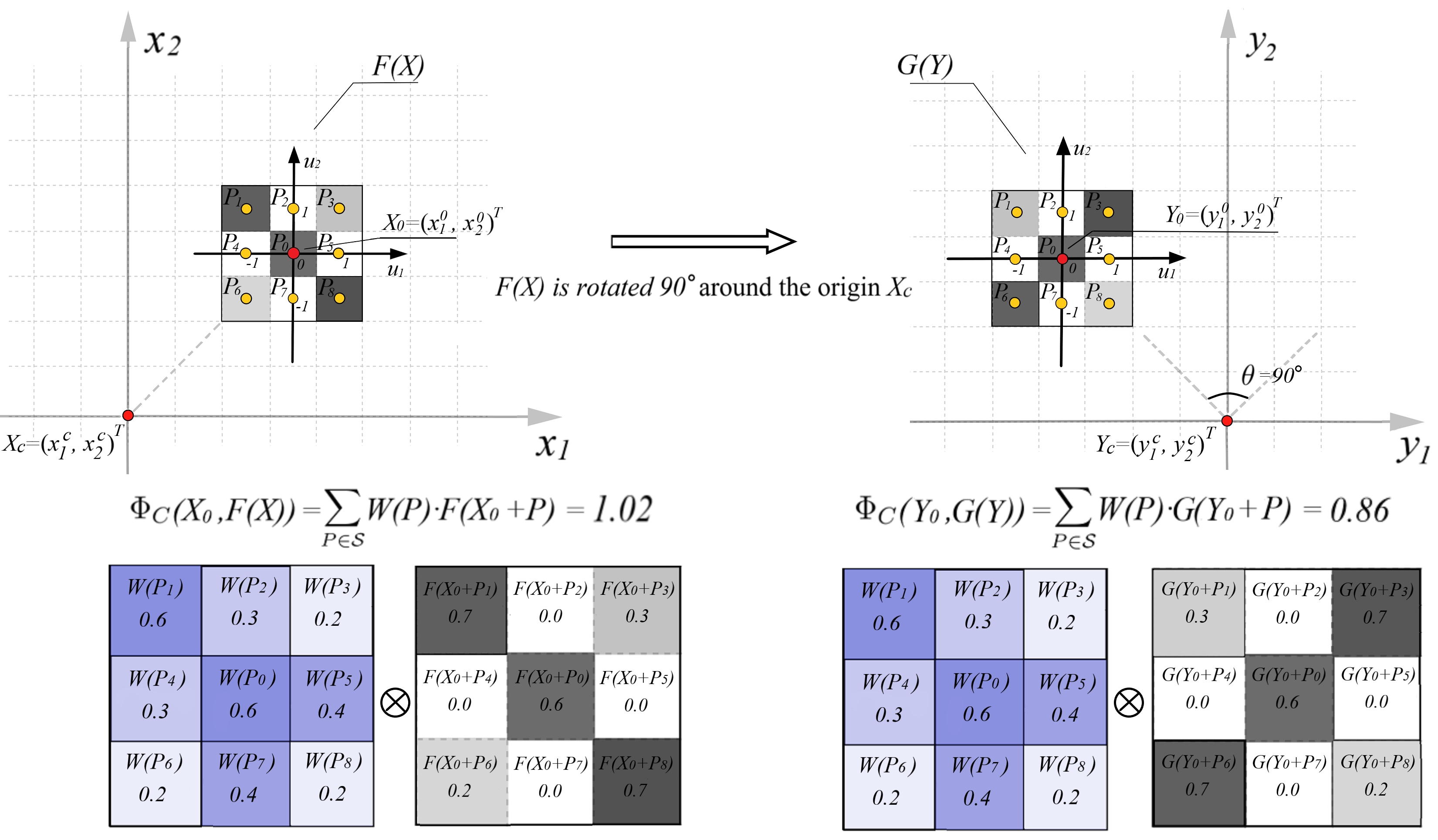}\label{figure:3(a)}\hfill}\\
	\subfloat[Illustration of the rotation invariance of $3\times3$ RIC-C.]
	{\includegraphics[height=60mm,width=110mm]{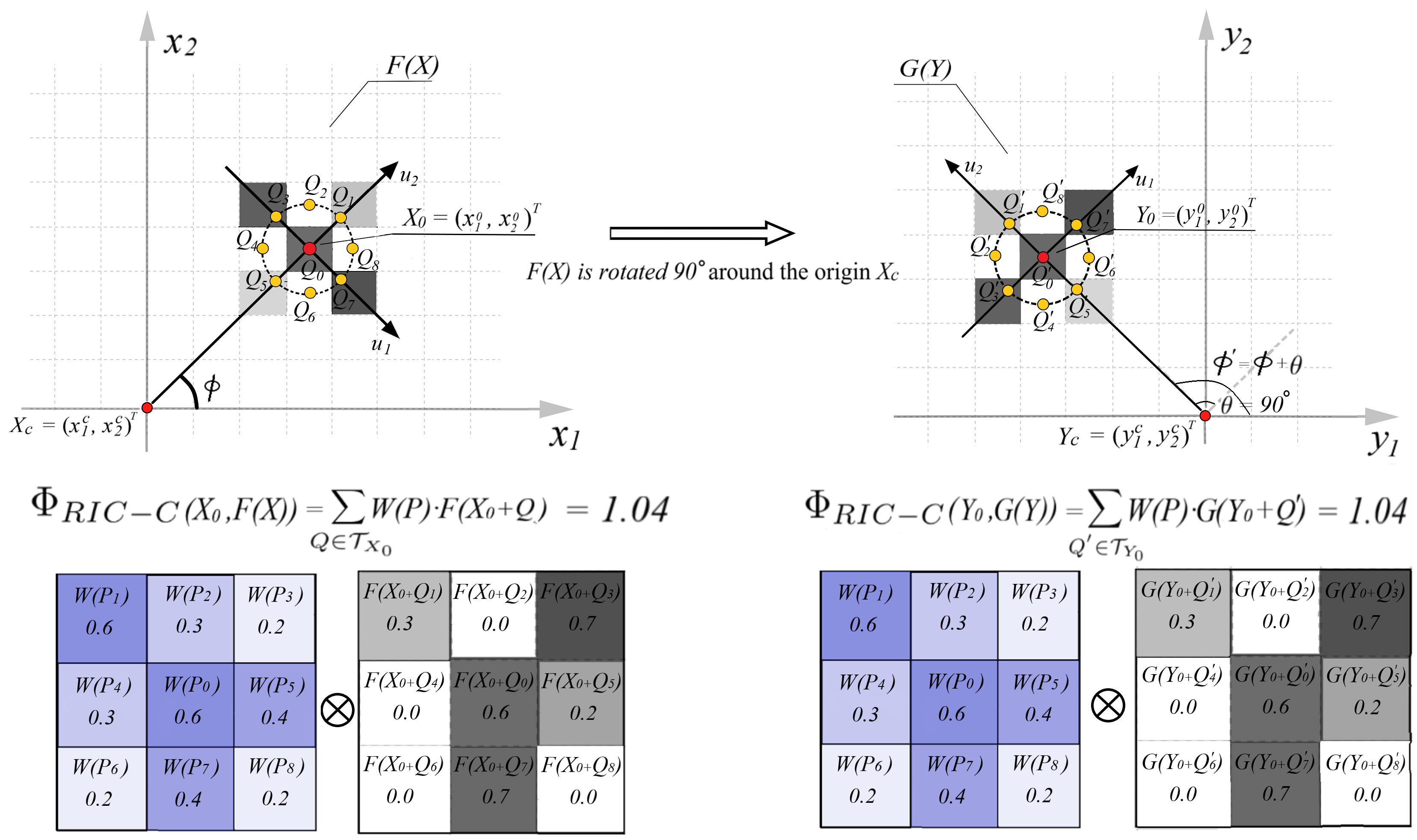}\label{figure:3(b)}\hfill}
	\caption{Suppose that the input $F(X)$ is rotated by $90^{\circ}$ around the center $X_{c}$ to $G(Y)$, and $Y_{0}\in\Omega^{'}$ is the corresponding point of $X_{0}\in\Omega$. Taking $3\times3$ convolutional kernel as example, we intuitively illustrate why $\Phi_{C}(X_{0},F(X))\neq\Phi_{C}(Y_{0},G(Y))$ and $\Phi_{RIC-C}(X_{0},F(X))=\Phi_{RIC-C}(Y_{0},G(Y))$. Note that the symbol $\otimes$ denotes a convolution operation.}\label{figure:3}
\end{figure}

Let $G(Y)$ be a rotated version of $F(X)$, and $Y_{0}\in \Omega^{'}$ is the corresponding point of $X_{0}\in\Omega$. Then, we have
\begin{equation}\label{equ:3}
	\Phi_{C}(Y_{0},G(Y))=\sum_{P\in\mathcal{S}}W(P)\cdot G(Y_{0}+P)
\end{equation}
According to $\left(\ref{equ:1}\right)$, the following relation can be obtained
\begin{equation}\label{equ:4}
	G(Y_{0}+P)=F(R_{-\theta}(Y_{0}+P))=F(R_{-\theta}Y_{0}+R_{-\theta}P)=F(X_{0}+R_{-\theta}P)\ne F(X_{0}+P)
\end{equation}
By substituting (\ref{equ:4}) into (\ref{equ:3}), we can get 
\begin{equation}\label{equ:5}
\Phi_{C}(Y_{0},G(Y))=\sum_{P\in\mathcal{S}}W(P)\cdot F(X_{0}+R_{-\theta}P)\ne\sum_{P\in\mathcal{S}}W(P)\cdot F(X_{0}+P)=\Phi_{C}(X_{0},F(X))
\end{equation}
Thus, the traditional convolutional operation $\Phi_{C}$ is not invariant to two-dimensional rotation. In Figure \ref{figure:3(a)}, we illustrate this with an intuitive example. 

\section{Related Work}
\label{section:3}
In this section, we briefly introduce the prior approaches to encode rotation invariance into image hand-crafted features and classical CNN models. In our opinion, most of them can be classified into the following four categories. 

\textbf{\emph{Orientation Assignment:}} In 1999, David proposed the image feature SIFT, which can be seen as a histogram of image gradient directions \cite{1}. It is a milestone in the fields of image processing and computer vision. Since then, researchers have designed a variety of SIFT-like features, including PCA-SIFT \cite{22}, SURF \cite{23}, DAISY \cite{24}, GLOH \cite{25} and so on. Since gradient directions are sensitive to image rotation, to ensure rotation invariance of these features, we need to first estimate a dominant orientation of a given image based on gradient directions and then rotate the image to a standardized orientation (the dominant orientation points upward). This process is called dominant orientation assignment. Inspired by this approach, in 2015, Jaderberg \emph{et al.} proposed the modified CNN model STN \cite{6}. They inserted spatial transformer module (ST), a tiny CNN, before each of convolutional layers of a classical CNN. Given an input (an image or a feature map), the ST was trained to learn a rotation angle and then transform the input into a canonical form before passing it to the next convolutional layer. Obviously, this method requires extra leanrable parameters to achieve rotation invariance, and we need to train ST with rotation augmentation. Meanwhile, recent study has proved that the ST module is not an option for achieving rotation invariance of intermediate feature maps \cite{26}. In fact, in most cases, traditional spatial transforms do not have ability to describe deformations of intermediate feature maps, because these deformations usually simultaneously act on spatial domain and vector domain. Hence, in theory, the ST can only be inserted before the first layer of a classical CNN. Recently, based on pixel-level gradient alignment operation before regular convolution, Hao et al. proposed Gradient-Aligned CNN (GA-CNN). In theory, this method does not increase learnable parameters and makes the model invariant to arbitrary rotations \cite{27}.       
         
\textbf{\emph{Polar/Log-Polar Transform:}} If two images are related by a rotation, the polar/log-polar image of the rotated image is translated along the vertical axis compared to the polar/log-polar image of the original image. Compared with image rotation, it is easier to handle image translation. Thus, some of rotation-invariant hand-crafted features are constrcuted based on polar/log-polar coordinate systems \cite{28,29}. Recently, some approaches also used log-polar representation of an image as the input of classical CNNs \cite{30,31,32}, because as stated above traditional convolution is natually invariant to image translation. However, converting coordinates not only causes information loss, but also changes the image size. 

\textbf{\emph{Multi-Orientation Feature Extraction:}} Inspired by biological vision systems, many researchers focused on designing various filters and filter banks to extract effective image features, such as Gabor filters, Gaussian derivatives and Wavelet filters. To achieve rotation invariance, these approaches usually apply multiple rotated versions of a filter to a given image, and then record only the maximum filter response across all orientations \cite{33,34,35}. In a CNN model, all convolutional operations can also be regarded as image filtering, which means we can use a similar way to handle image rotation. Based on the concept of symmetry group, Cohen \emph{et al.} proposed G-CNN \cite{10}. Roughly speaking, they filtered an input (an image or a feature map) using four rotated versions of a convolutional kernel (the rotation angle $\theta=k\cdot\pi/2$, where $k=0,1,2,3$), and then performed a max pooling within the four feature maps. Zhou \emph{et al.} proposed Oriented Response Networks (ORN) based on Active Rotating Filters (ARFs) \cite{36}. An ARF is a filter that actively rotates during convolution to produce a feature map with multiple orientation channels. In RotEqNet \cite{8}, each convolutional kernel is applied at multiple orientations and returns a vector field representing magnitude and angle of the highest scoring orientation at every spatial location. Previous studies proved that some image filters have steerability \cite{37,38}. A filter is rotationally steerable when its arbitrary rotated versions can be expressed as linear combinations of several basis filters. In order to reduce the number of learnable parameters and memory requirements, some papers constructed rotation-invariant CNNs based on various streeable filters, such as circular harmonic filters \cite{9,39}. Obviously, when applying a filter or a convolutional kernel to an input, rotating the filter or kernel is equivalent to rotating the input. Hence, there are also some approaches to obtain rotation invariance of CNNs by rotating inputs \cite{7,40,41}. For example, TI-Pooling first generated multiple rotated versions of an input image, and then employed the same CNN to extract the features of them respectively and performed a pooling operation across these feature \cite{7}. It is worth mentioning that most of multi-orientation convolutional operations are just invariant to several special rotation angles, and the number of rotated versions of a convolutional kernel (or an input) directly determines their invariance. In general, more rotated versions mean better performance but also mean higher computational cost.       

\textbf{\emph{Rotation-Invariant Coordinate System:}} Several studies found that orientation assignment is an error-prone process, thus result in the decline of SIFT-like features' performance \cite{42,43}. To resolve this problem, in 2005, Lazebnik \emph{et al.} designed a rotation-invariant coordinate system to calculate image gradient and then proposed a new image rotation-invariant feature RIFT \cite{3}. Using the same coordinate system, Fan \emph{et al.} and Wang \emph{et al.} further constructed several more effective rotation-invariant hand-crafted features than SIFT and RIFT, including Multi-Support Region Order-Based Gradient Histogram(MROGH) \cite{42}, LIOP \cite{4} and Overall Intensity Order Pattern(OIOP) \cite{44}. Based on the introduction above, we can find that researchers usually refer to the construction methods of hand-crafted features to modify classical CNN models. However, as far as we know, no previous research has made use of this coordinate system to encode rotational invariance into CNNs.   
  
\section{Rotation-Invariant Coordinate Convolutional Neural Network}
\label{section:4}
This section demonstrates how to perform a convolutional operation based on a rotation-invariant coordinate system. Then, we analyze the relationships and differences between RIC-C and deformable convolution (DEF-C) proposed in \cite{12}, and introduce a simple and effcient method to implement RIC-C.   

\subsection{Rotation-Invariant Coordinate Convolution}
\label{section:4.1}
In Section \ref{section:2.2}, we have proved that conventional convolution $\Phi_{C}$ is not invariant to two-dimensional rotation. To solve this problem, we design a new coordinate system $\mathcal{T}_{X_{0}}$ at a given point $X_{0}=\left(x^{0}_{1}, x^{0}_{2}\right)^{T}\in\Omega$. When setting a non-negative integer $n$, we also need to sample $\left(2n+1\right)\cdot\left(2n+1\right)=\left(4n^{2}+4n+1\right)$ points $Q$ on $\mathcal{T}_{X_{0}}$, and then calculate the sum of sampled values $F(X+Q)$ weighted by a $\left(2n+1\right)\times\left(2n+1\right)$ convolutional kernel $W$. 

Specifically, $8r$ sample points are equally distributed on a circle of radius $r$ centered at $X_{0}$, where $r=1,2,...,n$. To obtain a rotation invariant sampling, the first sample point is sampled along the radial direction which is from the origin $X_{c}$ to $X_{0}$. Since there are two sample points along the radial direction on a circle, the one which is farther from $X_{c}$ is selected as the first sample point. Then, the rest $\left(8r-1\right)$ points are sampled on the circle in an anticlockwise direction. Thus, besides $X_{0}$, other $(4n^{2}+4n)$ sample points $Q$ can be derived as follow 
\begin{equation}\label{equ:6}
Q = \left(r\cdot \cos{\left(\phi+\frac{i\cdot2\pi}{8r}\right)}, r\cdot \sin{\left(\phi+\frac{i\cdot2\pi}{8r}\right)}\right),~~i=0,1,...,\left(8r-1\right) ~~\mbox{and}~~r=1,...,n
\end{equation}
Note that the radial direction $\phi=atan2\left(x^{0}_{2}-x^{c}_{1}, x^{0}_{1}-x^{c}_{2}\right)\in(-\pi,\pi]$. Due to $X_{c}=\left(0,0\right)^{T}$, this formula can be further simplified as $\phi=atan2\left(x^{0}_{2}, x^{0}_{1}\right)$.     
  
Different from the regular grid $\mathcal{S}$ used in (\ref{equ:2}), the new coordinate system $\mathcal{T}_{X_{0}}$ is dependent on the position $X_{0}$ where $\Phi_{RIC-C}$ acting, because the angle $\phi$ is determined by $X_{0}$. Hence, different points $X_{0}$ have different $\mathcal{T}_{X_{0}}$. Figure \ref{figure:2(b)} shows how to order $\left(4n^{2}+4n+1\right)$ sample points $Q$ on $\mathcal{T}_{X_{0}}$, and the subscript of $Q_{\alpha}$ represents its order. Obviously, $Q_{\alpha}$ can be regarded as the corresponding point of $P_{\alpha}$ on the regular grid $\mathcal{S}$ (see Figure \ref{figure:2}).        

Then, RIC-C acting on $X_{0}$ can be defined as 
\begin{equation}\label{equ:7}
\Phi_{RIC-C}(X_{0},F(X))=\sum_{Q\in\mathcal{T}_{X_{0}}}W(P)\cdot F(X_{0}+Q)
\end{equation}
where $P\in\mathcal{S}$ has the same order as $Q$. We can prove that the convolutional operation based on the coordinate system $\mathcal{T}_{X_{0}}$ is invariant to any rotations around the image center. Suppose that $F(X)$ is rotated by $\theta$ degrees around $X_{c}$, and $G(Y)$ denotes this rotated version. Let the point $Y_{0}\in \Omega^{'}$ be the corresponding point of $X_{0}\in\Omega$, which means they are related by (\ref{equ:1}). Then, for each of $Q^{'}_{\alpha}$ in $\mathcal{T}_{Y_{0}}$, we have  
\begin{equation}\label{equ:8}
\begin{split}
Q^{'}_{\alpha}&=\left(r\cdot\cos{\left(\phi^{'}+\frac{i\cdot2\pi}{8r}\right)}, r\cdot \sin{\left(\phi^{'}+\frac{i\cdot2\pi}{8r}\right)}\right)\\&
=\left(r\cdot\cos{\left(\phi+\theta+\frac{i\cdot2\pi}{8r}\right)}, r\cdot \sin{\left(\phi+\theta+\frac{i\cdot2\pi}{8r}\right)}\right)\\&
=R_{\theta}Q_{\alpha}
\end{split}
\end{equation}
where $\phi^{'}=atan2\left(y^{0}_{2},y^{0}_{1}\right)$. Using (\ref{equ:1}) and (\ref{equ:8}), the following relation can be obtained
\begin{equation}\label{equ:9}
\begin{split}
G(Y_{0}+Q^{'})=F(R_{-\theta}(Y_{0}+Q^{'}))=F(R_{-\theta}Y_{0}+R_{-\theta}Q^{'})=F(X_{0}+Q)
\end{split}
\end{equation}
Based on (\ref{equ:9}), we can further get
\begin{equation}\label{equ:10}
\begin{split}
\Phi_{RIC-C}(Y_{0},G(Y))=\sum_{Q^{'}\in\mathcal{T}_{Y_{0}}}W(P)\cdot G(Y_{0}+Q^{'})=\sum_{Q\in\mathcal{T}_{X_{0}}}W(P)\cdot F(X_{0}+Q)
\end{split}
\end{equation} 
This means that $\Phi_{RIC-C}$ is invariant to arbitrary image rotations around the origin $X_{c}$. Taking $3\times 3$ RIC-C as an example, Figure \ref{figure:3(b)} explains this process more clearly. By comparing Figure \ref{figure:3(a)} and Figure \ref{figure:3(b)}, we can better understand the difference between traditional convolution and RIC-C. Both $\Phi_{C}$ and $\Phi_{RIC-C}$ have $\left(2n+1\right)\cdot\left(2n+1\right)$ learnable parameters. Hence, no extra parameters are used to achieve rotation invariance of RIC-C. 
 
By replacing all conventional convolutions in a CNN model with the corresponding RIC-C, we can get a RIC-CNN. As stated in Section \ref{section:2.1}, the rotation invariance of RIC-C ensures its rotation equivariance, which means the feature map $\Phi_{RIC-C}(Y, G(Y))$ can also be derived by rotating $\Phi_{RIC-C}(X, F(X))$ by $\theta$ degrees around the center $X_{c}$. Thus, no matter how many RIC-C layers are casaded in RIC-CNN, the output of final RIC-C layer still remains the rotation equivariance. Using max-pooling or average-pooling, we can make the final feature map with $1\times1$ spatial resolution. Then, this feature are invariant to arbitrary image rotations and can be input into other network structures, such as fully connected layers.      

\subsection{A Simple Implementation of RIC-C}
\label{section:4.2}
In previous sections, we suppose that $F(X)$ is a continous function. In practical, it is a discrete function, whose domain $\Omega$ is a $H \times H$ regular grid. In our paper, we demand $H$ is a positive even integer, and the center $X_{c}=\left(\left(H+1\right)/2,\left(H+1\right)/2\right)$. Since $X_{c}$ is fractional and does not lie on $\Omega$, this setting ensures that $X_{c}$ and any $X_{0}\in\Omega$ are not coincident, and allows us to calculate the radial direction $\phi$ for each of $X_{0}$. 

\begin{figure}
	\centering
	\subfloat[The feature map computed using $3\times3$ DEF-C.]
	{\includegraphics[height=35mm,width=63mm]{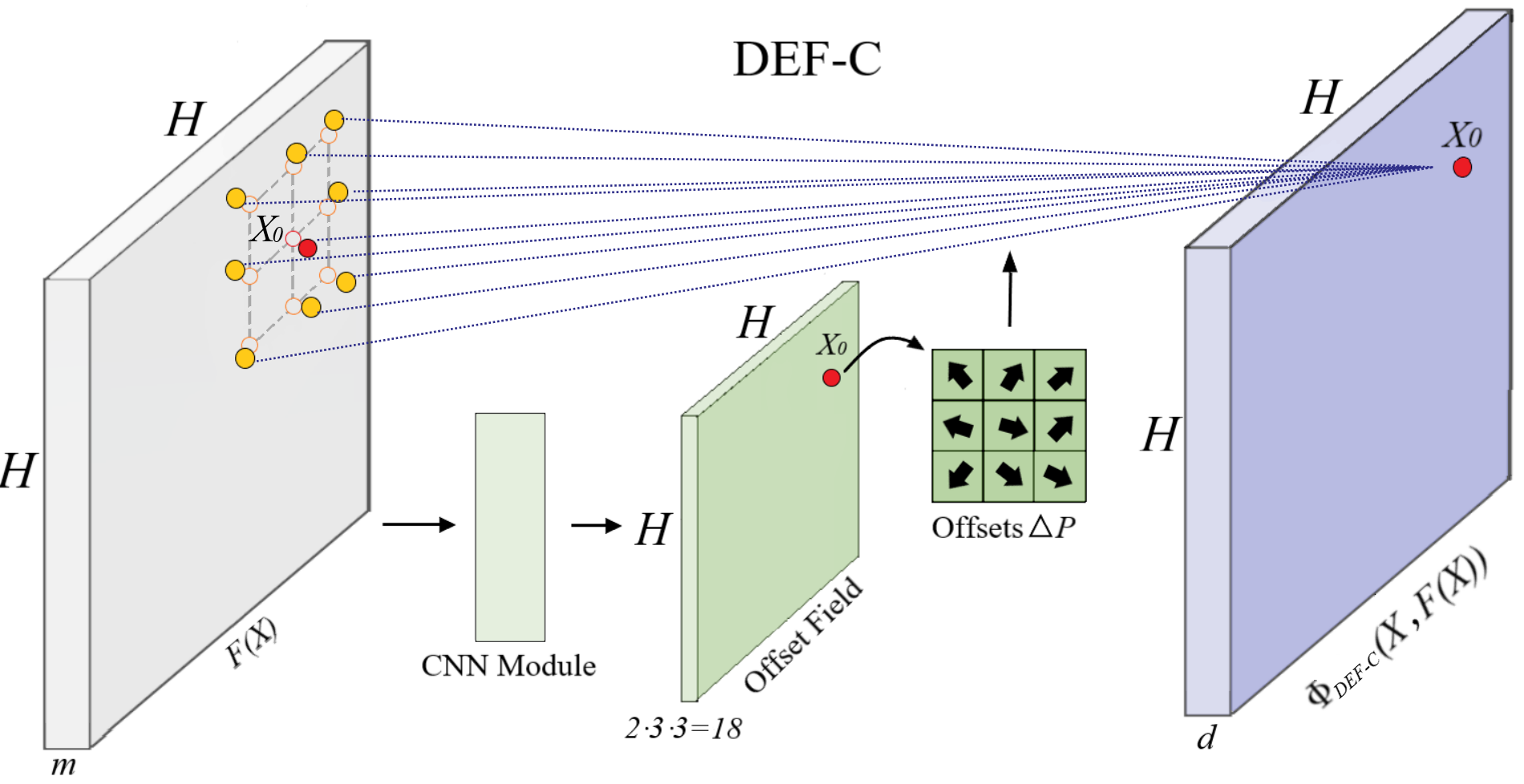}\label{figure:4(a)}\hfill}~~~~~
	\subfloat[The feature map computed using $3\times3$ RIC-C.]
	{\includegraphics[height=35mm,width=63mm]{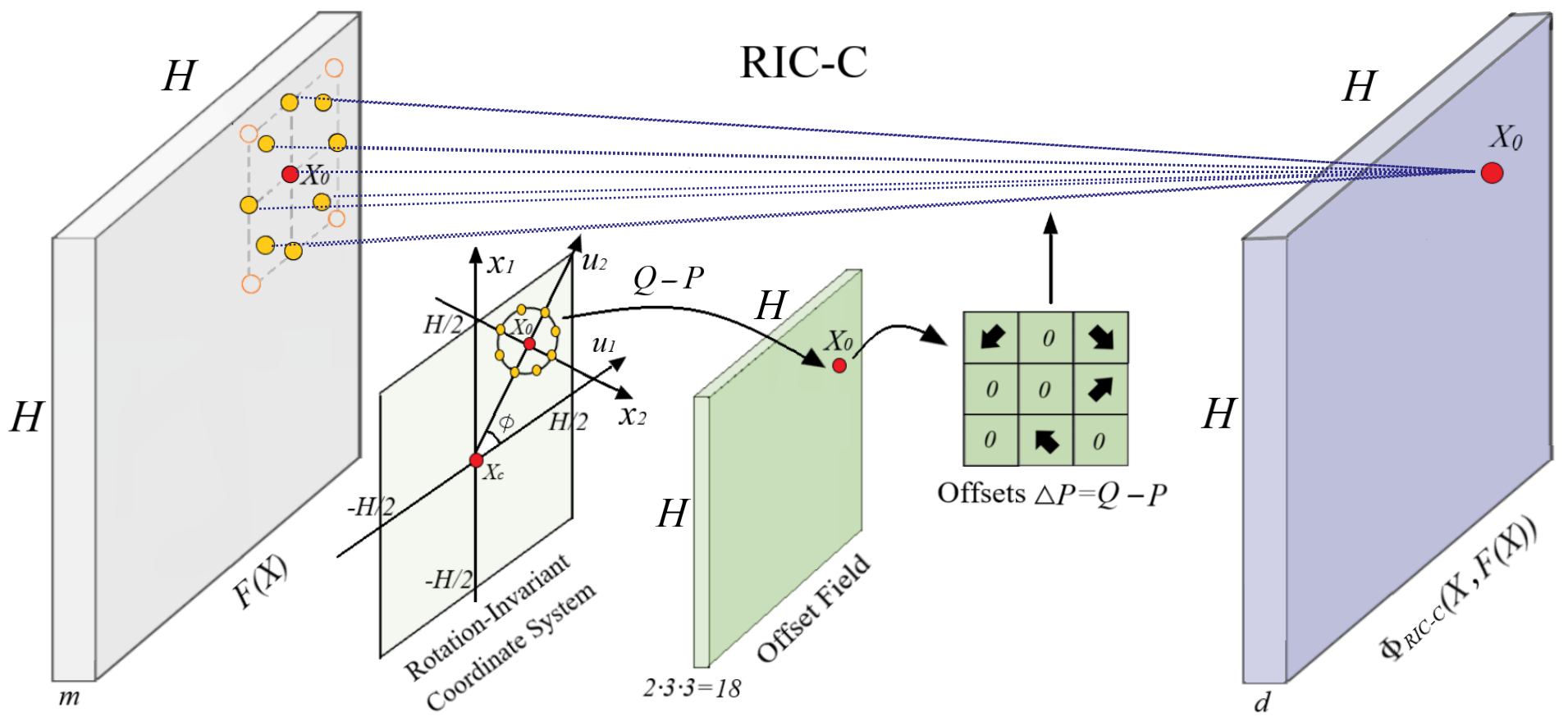}\label{figure:4(b)}\hfill}
	\caption{The connection and difference between $3\times 3$ DEF-C and $3\times 3$ RIC-C. Note that the offset field has the same spatial resolution with the input $F(X)$, and its channel dimension $18$ corresponds to $3\times3=9$ two-dimensional offsets.}\label{figure:4}
\end{figure}

Typically, the sample points $Q\in\mathcal{T}_{X_{0}}$ are also fractional. This means we have to calculate $F(X_{0}+Q)$ via bilinear interpolation. In 2017, Dai et al. proposed deformable convolutional networks \cite{12}. Given a non-negative integer $n$, a $\left(2n+1\right)\times\left(2n+1\right)$ DEF-C can be defined as follow 
\begin{equation}\label{equ:11}
\Phi_{DEC-C}(X_{0},F(X))=\sum_{P\in\mathcal{S}}W(P)\cdot F(X_{0}+P+\Delta P)
\end{equation}
where the offset $\Delta P$ is learnable parameter, which is obtained by applying a convolutional module over the input $F(X)$. We find RIC-C defined by (\ref{equ:7}) can be expressed in a similar fashion 
\begin{equation}\label{equ:12}
\begin{split}
\Phi_{RIC-C}(X_{0},F(X))=\sum_{P\in\mathcal{S}}W(P)\cdot F(X_{0}+P+(Q-P))
\end{split}
\end{equation}
Now, $\Delta P$ is equal to the offset between $P$ and the corresponding $Q\in \mathcal{T}_{X_{0}}$ with the same order. Compared with DEF-C, the offsets in RIC-C are constants independent of the input content. Their values are predetermined by our rotation-invariant coordinate system without training, which ensure RIC-C's invariance to image rotation. When setting $n=1$, Figure \ref{figure:4} illustrates how the feature maps are calculated using $3\times 3$ DEF-C and $3\times 3$, respectively, and shows the connection and difference between them more clearly. In Pytorch, the function \textit{DEFORM\_CONV2D}\footnote[1]{\url{https://pytorch.org/vision/main/generated/torchvision.ops.deform_conv2d.html}} performs DEF-C. Thus, we can use it to implment RIC-C by just replacing $C\times H\times H$ learnable parameters of the offset field with the corresponding constants calculated based on the rotation-invariant coordinate system, where the channel dimension $C=2\cdot\left(2n+1\right)\cdot \left(2n+1\right)$. In next section, we will show that this simple implementation of RIC-C is very efficient.     

\section{Experiments}
\label{section:5}
In this section, extensive experiments are conducted on various datasets to evaluate the effectiveness of RIC-CNN. Using MNIST dataset \cite{13}, we train RIC-CNN on the original training set without data augmentation, and then validate its performance on the rotated test set. Some previous rotation-invariant CNN models are also chosen for comparison. Then, RIC-C is deployed to several classical CNN models, and real image classification is carried out on NWPU VHR-10 and MTARSI datasets \cite{17,18}. Finally, using triplet loss, we train a shallow CNN and the corresponding RIC-CNN on UBC benchmark dataset \cite{20}, and compare their performance on patch verification. Our models are implemented with Pytorch, and we use a NVIDIA P100 GPU for training and test.

\subsection{The Rotation Invariance of RIC-CNN}
\label{section:5.1}
\begin{figure}
	\centering
	\subfloat[Some samples from the training set.]
	{\includegraphics[height=30mm,width=50mm]{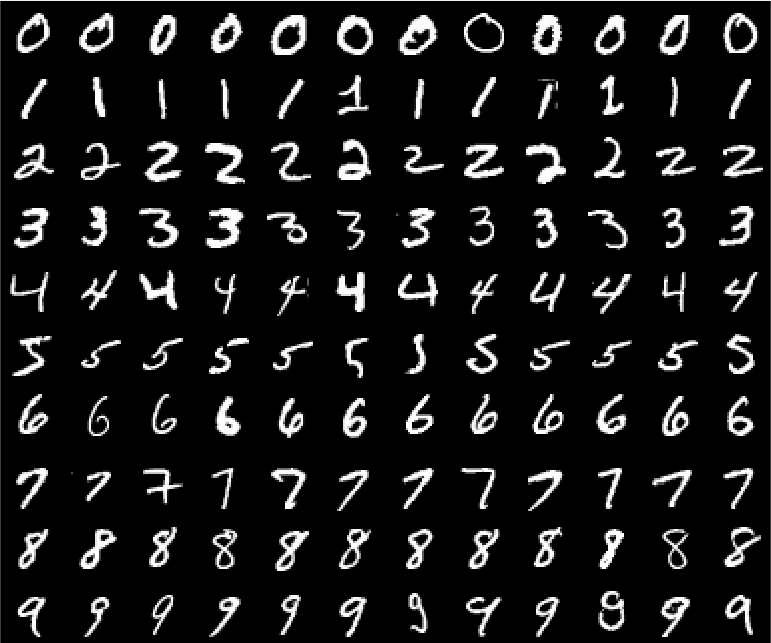}\label{figure:5(a)}\hfill}~~~~~
	\subfloat[Some samples from the rotated test set.]
	{\includegraphics[height=30mm,width=50mm]{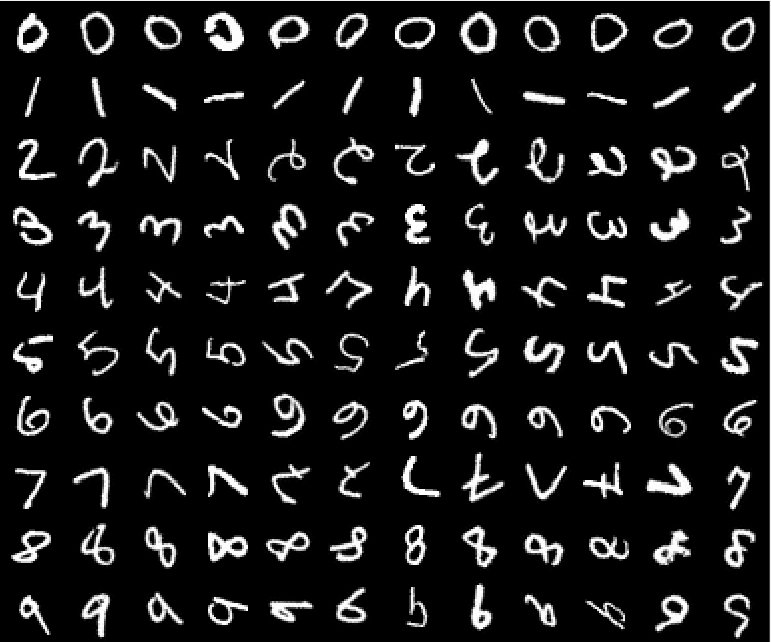}\label{figure:5(b)}\hfill}
	\caption{The training set and rotated test set of MNIST dataset.}\label{figure:5}
\end{figure}

We first evaluate the rotation invariance of RIC-CNN based on MNIST dataset \cite{13}. This dataset was collected for recognizing 10 handwritten digits (0$\sim$9), which contains 60K training images and 10K test images. We select 10K images from the training set for validation, and all images are normalized to $32\times32$ pixels. Then, each of test images is rotated around its center by angles from $0^{\circ}$ to $360^{\circ}$ every $10^{\circ}$. As a result, we derive 36$\times$10K=360K rotated test images, and they constitute our rotated test set. Some examples in the training set and rotated test set are shown in Figure \ref{figure:5}. Unlike previous papers \cite{9,10,36}, we do not use MNIST-rot-12k dataset \cite{45} to evaluate rotation-invariant CNNs, because both training and test images in that dataset are randomly rotated by angles in $[0,2\pi)$. As mentioned previously, this makes it difficult to verify whether new network architecture or just rotated training images enhances a model's rotation invariance. In fact, on that dataset, conventional CNNs can obtain comparable results with rotation-invariant CNNs.      
  
A standard CNN model is used as our baseline. It is made of six convolutional layers with 32, 32, 64, 64, 128 and 128 kernels of size $3\times3$, respectively; one $2\times2$ max pooling operation after the second/fourth convolutional layer; one $8\times8$ average pooling after the last layer; and one fully connected layer with 10 units. We apply batch normalization after convolutional layers and before activation functions ReLU. By replacing all $3\times3$ traditional convolutions in this baseline with $3\times3$ DEF-C and $3\times3$ RIC-C, we obtain the corresponding DEF-CNN and RIC-CNN, respectively. For DEF-CNN, an additional convolutional layer with 18 kernels of size $3\times3$ is inserted before each DEF-C layer to learn the offset field. In addition, it should be noted that $8\times8$ average pooling makes the feature map with $1\times1$ spatial resolution, and then this feature vector is input to the fully connected layer. As stated in Section \ref{section:4.1}, this ensures that RIC-CNN is theoretically invariant to arbitrary rotations. We use the same protocol to train these three models. Specifically, the number of epochs and the batch size are 100. For the optimization, the Adam optimizer is selected while the initial learning rate is $10^{-4}$ and it is multiplied by $0.8$ every 10 epochs. Also, seven existing rotation-invariant CNNs are chosen for comparison, including ORN\cite{36}, RotEqNet\cite{8}, G-CNN\cite{10}, H-Net\cite{9}, GA-CNN\cite{27}, B-CNN\cite{46} and E(2)-CNN\cite{47}. When data augmentation is not used in training, H-Net, GA-CNN, B-CNN and E(2)-CNN have the invariance to continous/arbitrary rotation angles, while other models are only invariant to some special angles (ORN: $k\cdot\pi/8, k=0,1,...,7$; RotEqNet: $k\cdot\pi/17, k=0,1,...,16$; G-CNN: $k\cdot\pi/2, k=0,1,2,3$). These rotation-invariant CNN models are trained by using the codes and protocols provided by the authors. We do not choose those methods that completely dependent on data augmentation to obtain rotation invariance, such as STN\cite{6} and TI-Pooling\cite{7}.      

\begin{table}
	\caption{\label{table:1} The classification accuracies from various methods on the original test set and rotated test sets of MNIST. Underline and bold stand for suboptimal and best results.}
	\centering
	\begin{tabular}{p{2.8cm}p{2.5cm}p{2.5cm}p{2.5cm}}
		\toprule[1.3pt]
		Methods & Input Size & Original Test Set & Rotated Test Set\\
		\toprule[1.1pt]
		ORN\cite{36} & 32$\times$32 & 99.42\% & 80.01\%\\
		RotEqNet\cite{8} & 28$\times$28 & 99.26\% & 73.20\% \\
		G-CNN\cite{10} & 28$\times$28 & 99.27\% & 44.81\% \\
		H-Net\cite{9} & 32$\times$32 & 99.19\% & 92.44\% \\
		GA-CNN\cite{27} & 28$\times$28 & 95.67\% & 93.29\% \\
		B-CNN\cite{46} & 28$\times$28 & 97.40\% & 88.29\% \\
		E(2)-CNN\cite{47} & 29$\times$29 & 98.14\% & \underline{94.37}\% \\
		\midrule
		CNN & 32$\times$32 & \underline{99.55}\% & 45.42\% \\
		DEF-CNN\cite{12} & 32$\times$32 & \textbf{99.67\%} & 46.97\% \\
		RIC-CNN & 32$\times$32 & 99.02\% & \textbf{95.52\%} \\
		\midrule
	\end{tabular}
\end{table} 

\begin{figure}
	\centering
	\includegraphics[width=90mm,height=45mm]{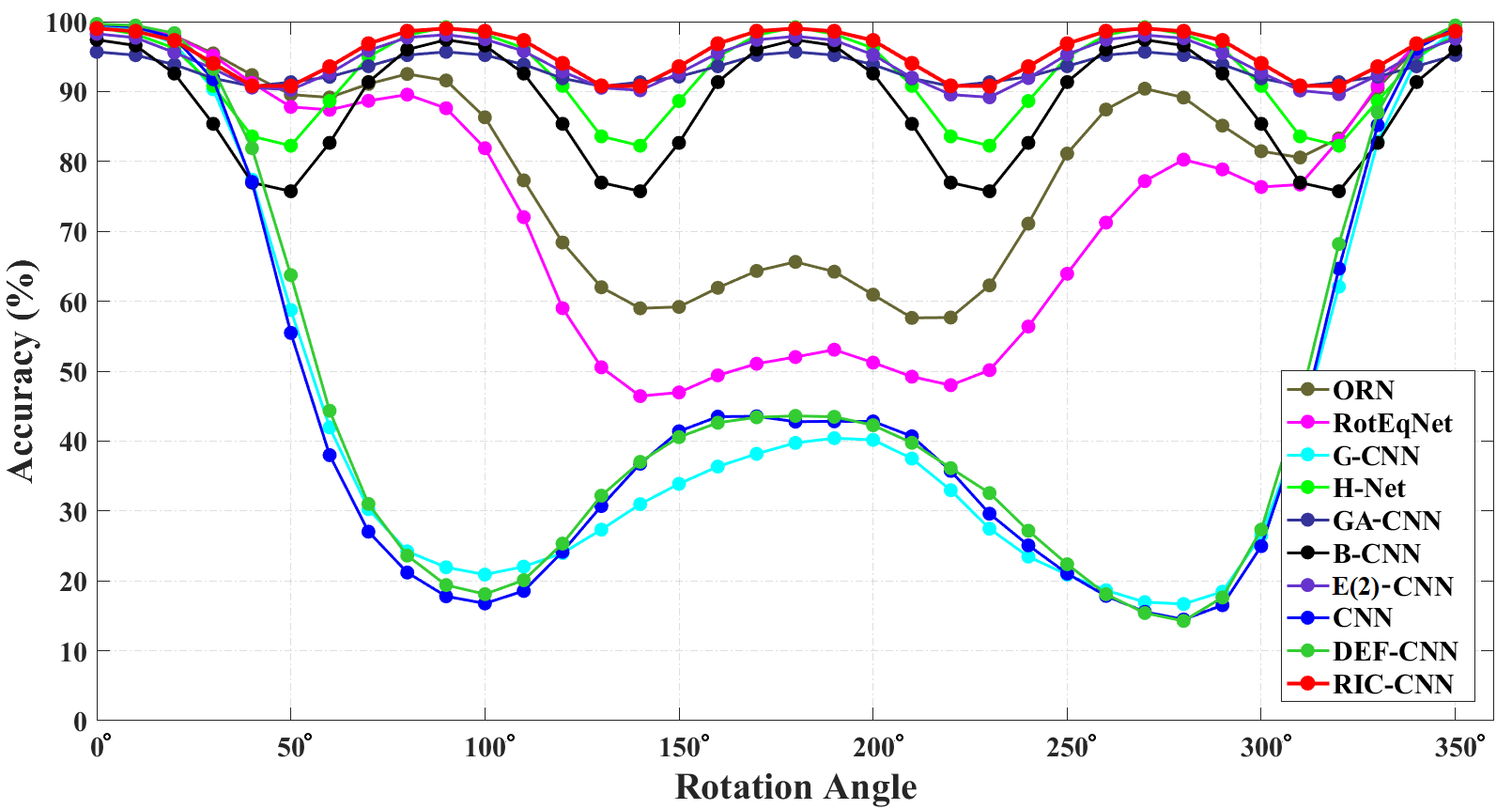}\\
	\caption{The classification accuracies from using various methods on 36 rotated test subsets of MNIST with specific rotation angles (0, $10^{\circ}$, $20^{\circ}$,...,$350^{\circ}$).}\label{figure:6}
\end{figure}

Table \ref{table:1} lists the classification accuracies from using various methods on the original test set and rotated test set of MNIST. First, we can see that RIC-CNN achieves the best accuracy rate (95.52\%) on the rotated test set, and significantly outperforms traditional CNN (45.42\%) and DEF-CNN (46.97\%). Without data augmentation, DEF-CNN also has poor performance, because it is unable to learn suitable offset field from training data to achieve rotation invariance. The second observation is that on the rotated test set, the five models with the invariance to arbitrary rotations perform better than the three models that are only invariant to several special angles. A surprising finding is that G-CNN performs as bad as standard CNN. This suggests that in addition to group convolutions, G-CNN also requires data augmentation to learn rotation invariance. Furthermore, we divide the rotated test set into 36 subsets, and each of subsets contains all 10K test images with the same rotation angle (0, $10^{\circ}$, $20^{\circ}$,...,$350^{\circ}$). Figure \ref{figure:6} shows the accuracy rates from different models on these 36 subsets. It can be seen that RIC-CNN achieves the best accuracy on most of test subsets with different rotation angles, and E(2)-CNN performs almost as well as RIC-CNN. In addition, the accuracy curves of H-Net, GA-CNN, B-CNN, E(2)-CNN and RIC-CNN have a period of $90^{\circ}$. In fact, pooling operations, such as max pooling, are not strictly rotation invariant. When the rotation angle is not a multiple of $90^{\circ}$, both pooling and interpolation operations will introduce calculation errors and reduce the invariance of these models. As the rotation angle increases from 0 to $90^{\circ}$, this error first increases and then decreases. Thus, we can observe that in a period, the accuracy first decreases ($0^{\circ}\sim 45^{\circ}$) and then increases ($45^{\circ}\sim 90^{\circ}$). It is also obvious that the rotation invariance of H-Net and B-CNN is worse than that of RIC-CNN and E(2)-CNN, when the rotation angle is around $45^{\circ}$. This indicates that the calculation errors have a greater impact on their performance. 

\begin{figure}
	\centering
	\subfloat[Rotated versions of some digits are very similar with digits from other classes. Without orientation information, it is difficult to classify these samples correctly.]
	{\includegraphics[height=35mm,width=60mm]{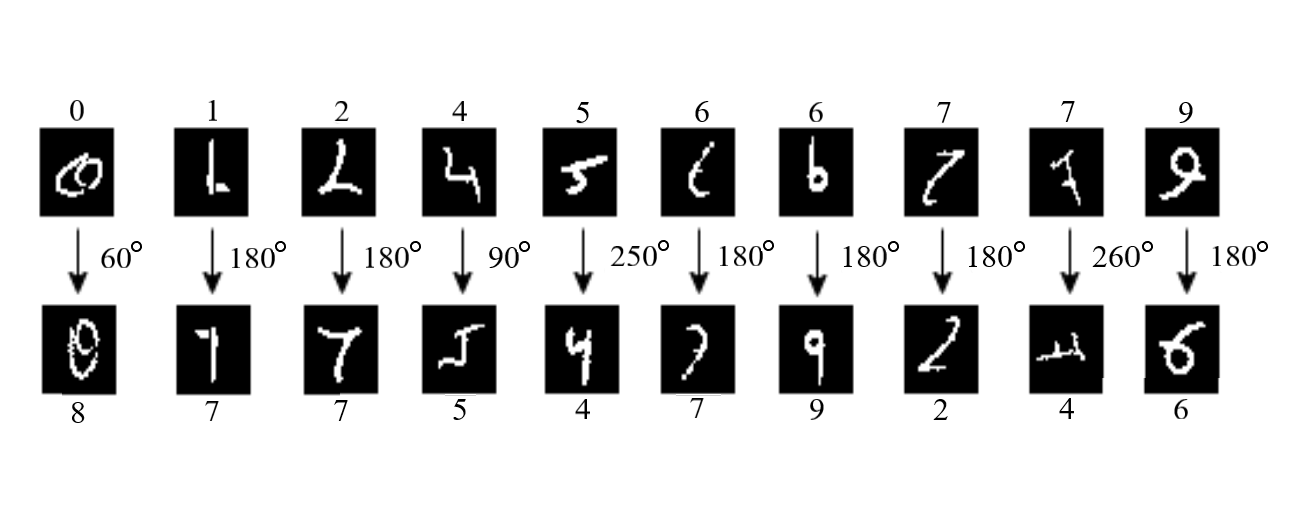}\label{figure:7(a)}\hfill}~~~~~
	\subfloat[The classification accuracies from using conventional CNN and RIC-CNN on each of digit classes.]
	{\includegraphics[height=35mm,width=60mm]{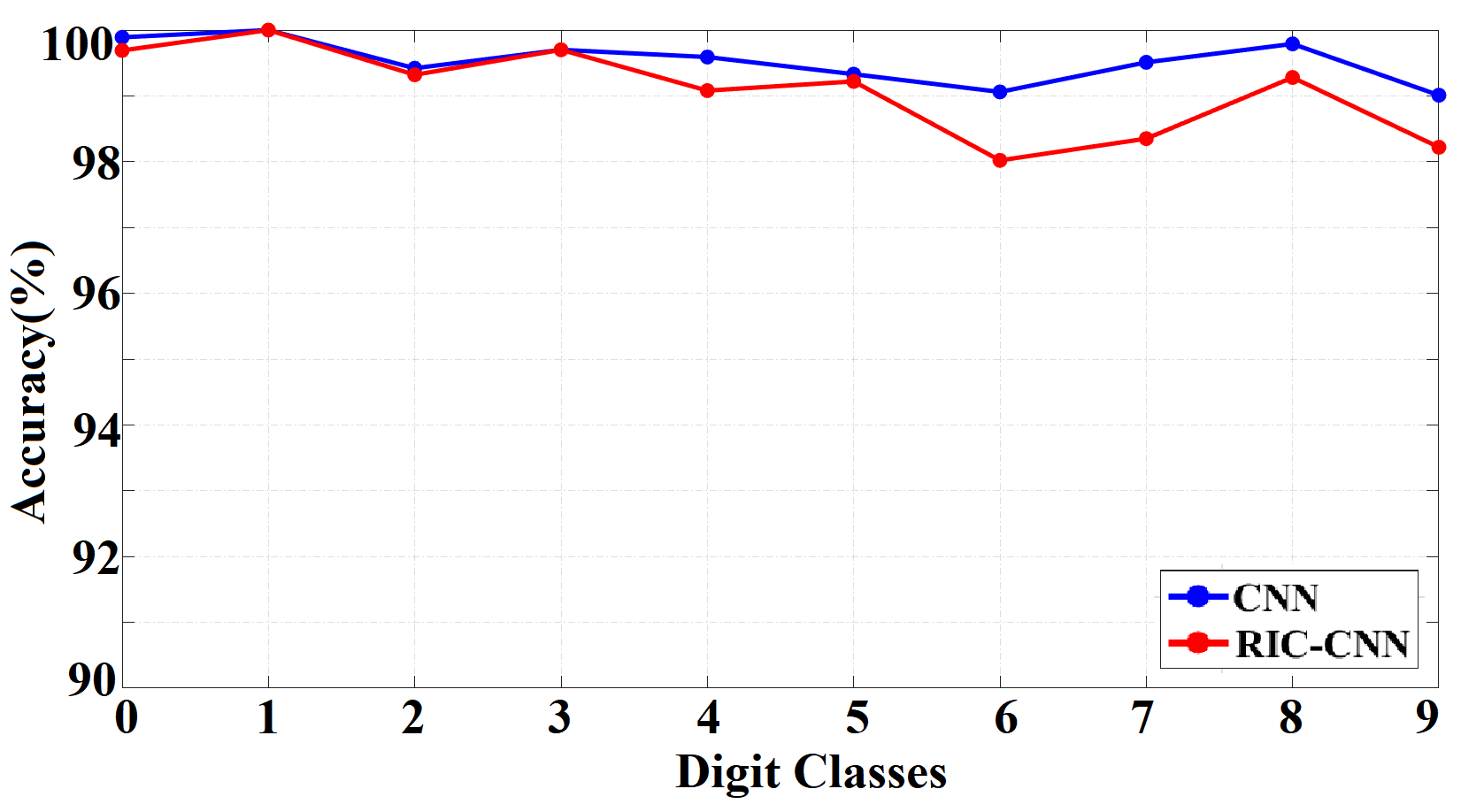}\label{figure:7(b)}\hfill}\\
	\caption{The reason why rotation-invariant CNNs perform worse than conventional CNN on the original test set of MNIST.}\label{figure:7}
\end{figure}

In addition, when a CNN model has the invariance to continous rotations, its performance on the original test set of MNIST will decline. For example, the accuracy rates from using H-Net, GA-CNN, B-CNN, E(2)-CNN and RIC-CNN are 99.19\%, 95.67\%, 97.40\%, 98.14\% and 99.02\%, respectively. Clearly, they are worse than conventional CNN (99.55\%), DEF-CNN (99.67\%) and those models with the invariance to some special rotation angles. In fact, this is because rotated versions of some handwritten digits are very similar with digits from other classes, such as a digit 9 and a digit 6 rotated by $180^{\circ}$. Figure \ref{figure:7(a)} shows some ambiguous examples. Traditional CNNs can correctly classify these samples, but they are misidentified by rotation-invariant CNNs. In fact, the orientation information is crucial to correctly predict the classes of these samples. However, rotation-invariant models have filtered out this information when extracting features. As shown in Figure \ref{figure:7(b)}, we also calculate the accuracy rates from using standard CNN and RIC-CNN on each of digit classes. It can be seen that RIC-CNN maintains high accuracies on digits 0$\sim$3 and 5, but its performance notablely drops on digits 4 and 6$\sim$9, especially on 6 and 9. Even so, RIC-CNN still outperforms GA-CNN, B-CNN and E(2)-CNN on the original test set of MNIST.        

\begin{figure}
	\centering
	\subfloat[The MRE curves of different RIC-C layers in RIC-CNN.]
	{\includegraphics[height=30mm,width=60mm]{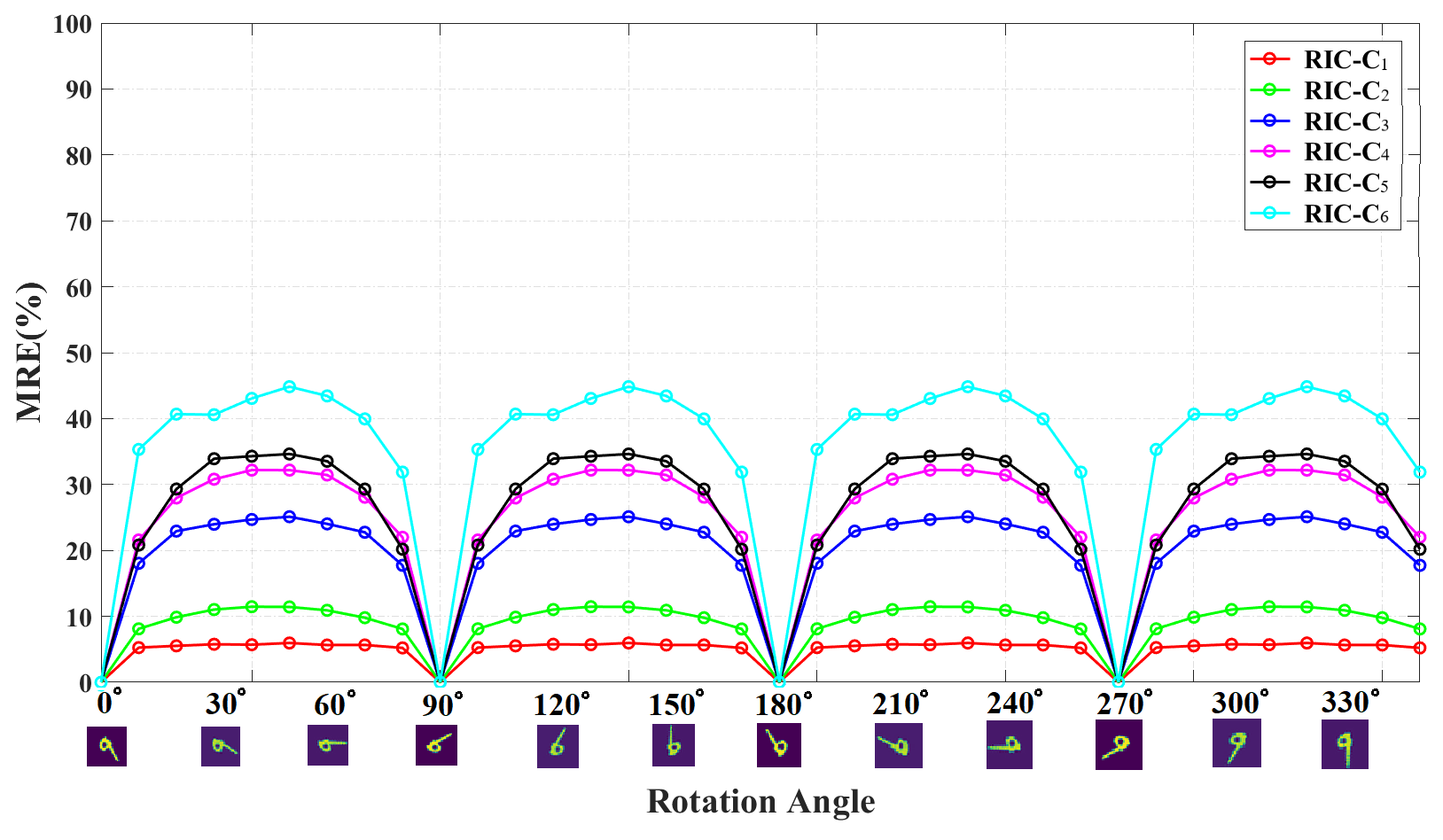}\label{figure:8(a)}\hfill}~~~~~
	\subfloat[The MRE curves of different convolutional layers in CNN.]
	{\includegraphics[height=30mm,width=60mm]{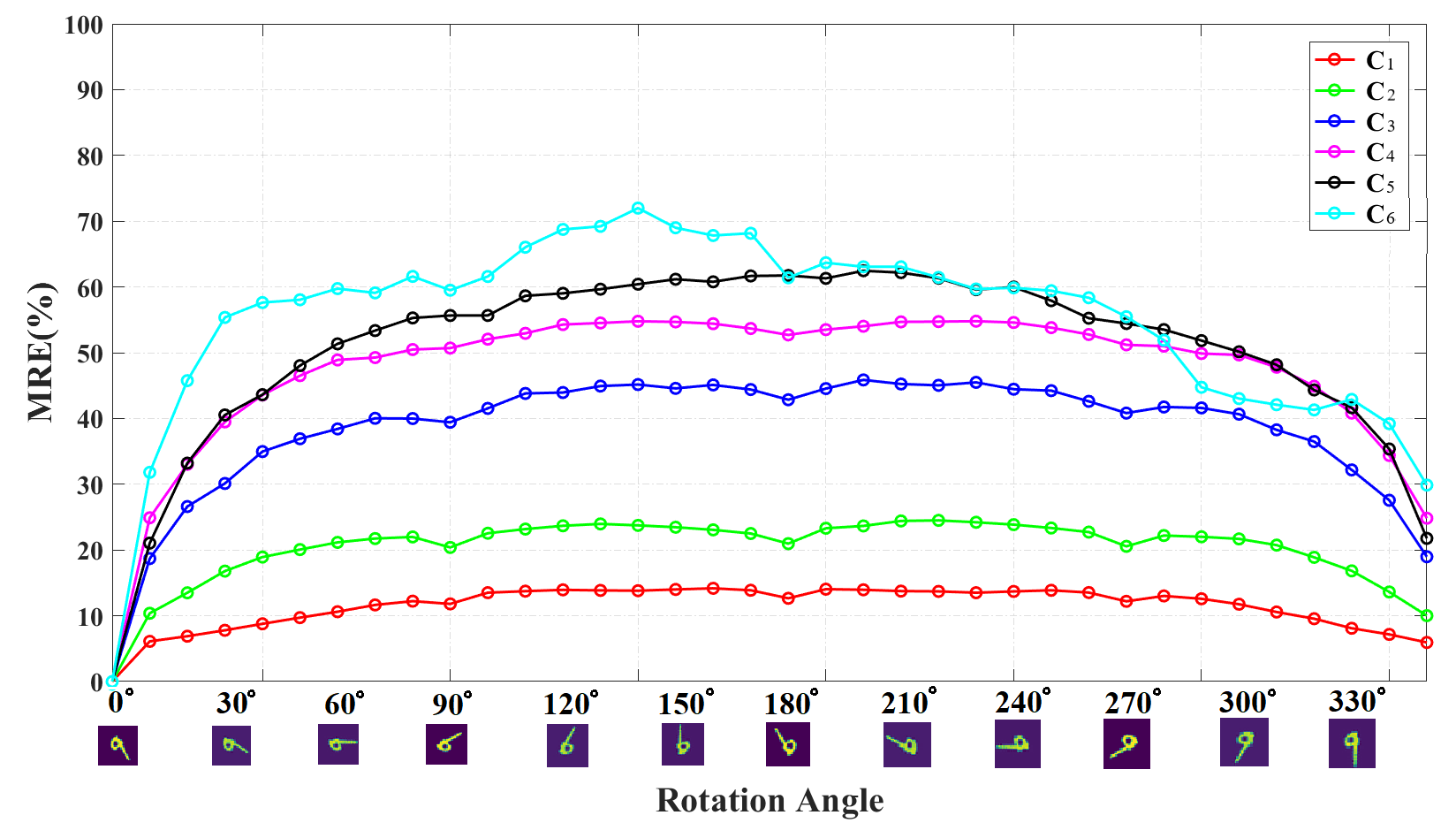}\label{figure:8(b)}\hfill}\\
	\caption{Illustration of the rotation equivariance of feature maps in RIC-CNN.}\label{figure:8}
\end{figure}

\begin{figure}
	\centering
	\subfloat[The feature maps of six RIC-C layers.]
	{\includegraphics[height=40mm,width=60mm]{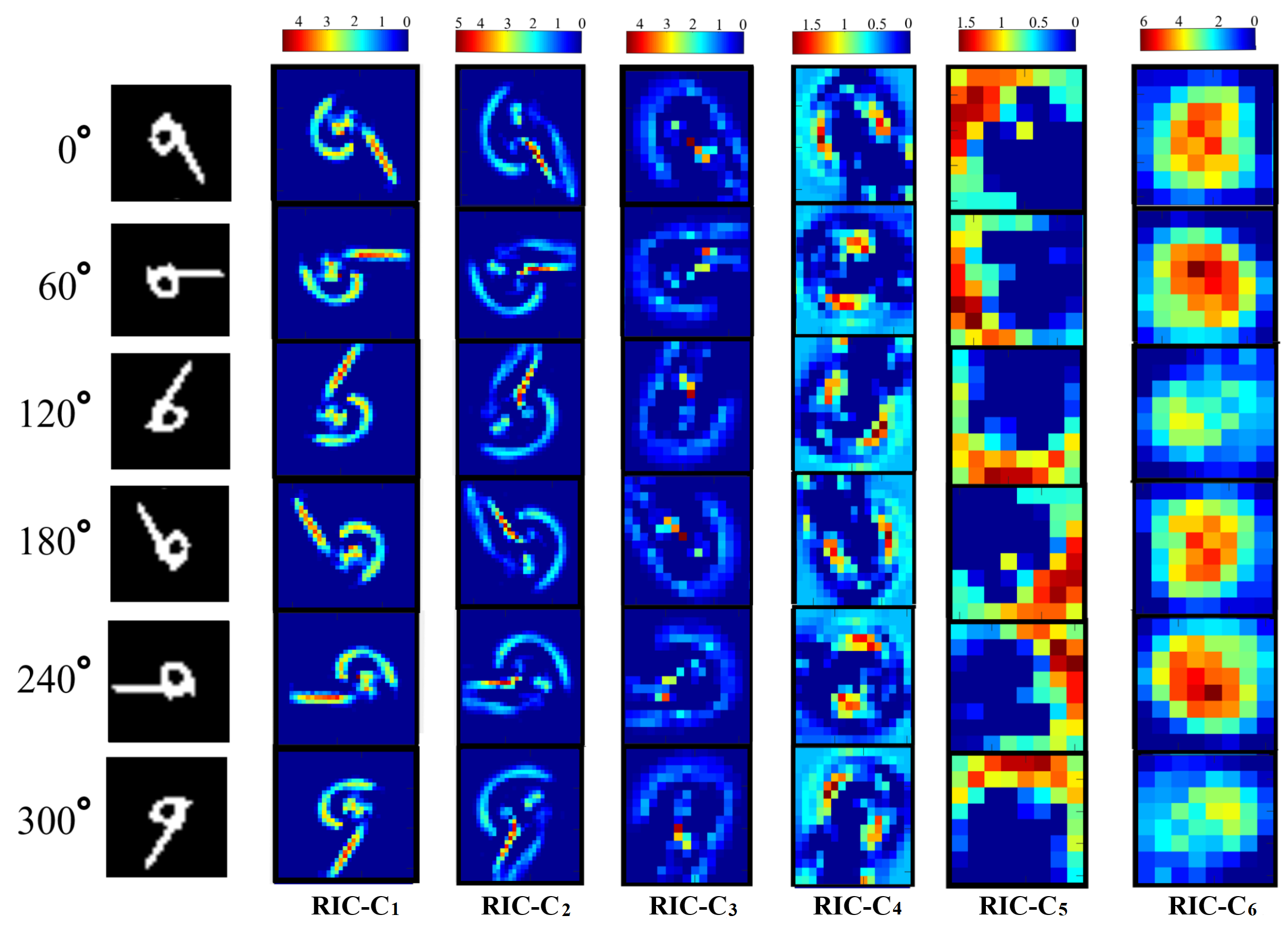}\label{figure:9(a)}\hfill}~~~~~
	\subfloat[The feature maps of six convolutional layers.]
	{\includegraphics[height=40mm,width=60mm]{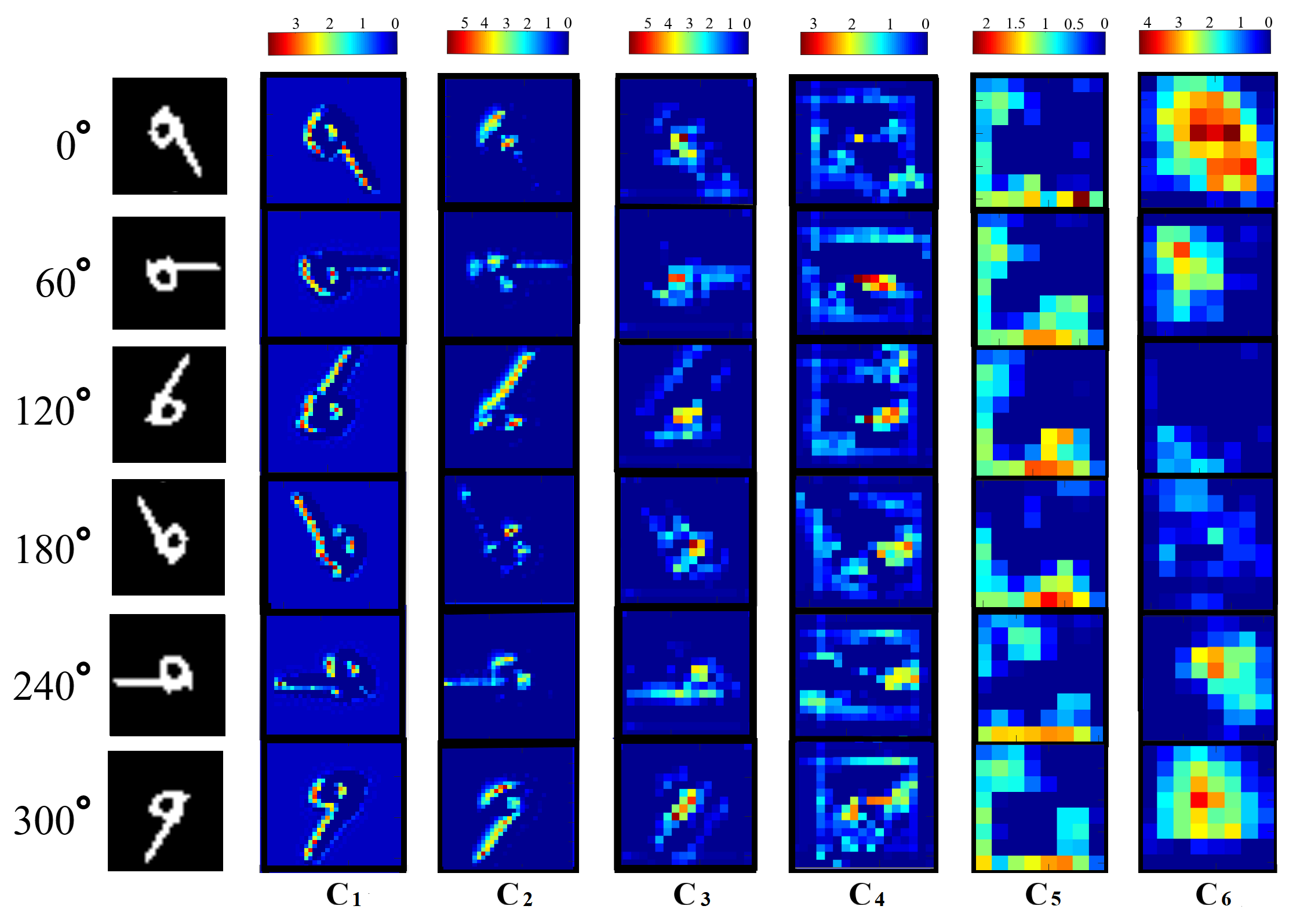}\label{figure:9(b)}\hfill}\\
	\caption{Visualization of feature maps of different layers in RIC-CNN and standard CNN. }\label{figure:9}
\end{figure}

As mentioned in Section \ref{section:2.1} and \ref{section:4.1}, the rotation invariance of RIC-C ensures that its feature map has rotation equivariance. To verify this, we first select an image $F(X)$ and its 35 rotated versions $G_{a}(Y)$ from the rotated test set, where $a=1,2,...,35$ and $G_{a}(Y)$ is derived by rotating $F(X)$ by $10\cdot a$ degrees around the center. Then, for $F(X)$ and each of $G_{a}(Y)$, the feature maps of six RIC-C layers in RIC-CNN are calculated, respectively. They are denoted as $\Phi_{RIC-C_{l}}(X,F(X))$ and $\Phi_{RIC-C_{l}}(Y,G_{a}(Y))$, where $RIC-C_{l}$ represents the $l^{th}$ RIC-C layer in RIC-CNN and $l=1,2,...,6$. In theory, if we rotate $\Phi_{RIC-C_{l}}(Y,G_{a}(Y))$ by $-10\cdot a$ degrees, it should be exactly the same as $\Phi_{RIC-C_{l}}(X,F(X))$. Thus, we do this and compute the relative error between the values of the rotated feature map and $\Phi_{RIC-C_{l}}(X,F(X))$ at the same spatial position and on the same channel. For two real numbers $x$ and $y$, the relative error between them is defined as $\left|x-y\right|/\left(\left|x\right|+\left|y\right|\right)\times 100\%$, where the operation $\left|\cdot\right|$ returns the absolute values of a given number. After that, the mean value of the relative errors over the whole feature map is calculated, which is denoted as $MRE^{l}_{a}$ ($0\leq MRE^{l}_{a}\leq 100\%$). Obviously, for each of $RIC-C_{l}$ layers, we finally derive a MRE curve which contain 35 sample points $MRE^{l}_{a}$ corresponding to different rotation angles. All of six MRE curves are shown in Figure \ref{figure:8(a)}. Using the same process, we also compute the MRE curves of six convolutional layers in CNN (these standard convolutions are denoted as $C_{1},C_{2},...,C_{6}$), and show them in Figure \ref{figure:8(b)}. The first observation is that the MRE curves of different RIC-C layers all have a period of $90^{\circ}$. In a period, with the increase of the rotation angle, the error first increases ($0^{\circ}\sim 45^{\circ}$) and then decreases ($45^{\circ}\sim 90^{\circ}$). Similar to Figure \ref{figure:6}, this phenomenon is also caused by interpolation and max pooling operations. In fact, for any $l$, when $a=90^{\circ},180^{\circ},270^{\circ}$, we always have $MRE^{l}_{a}\approx 0$, which means that all feature maps have perfect rotation equivariance to these rotation angles. Secondly, for any $a$, $MRE^{l}_{a}$ increases as $l$ increases. In Figure \ref{figure:8(a)}, the maximum value of $MRE^{1}_{a}$ is only 5.96\%, which means the feature map of the first RIC-C layer keeps good rotation equivariance. However, the maximum MRE of $RIC-C_{6}$ increases to 44.83\%. This is because the errors from the previous layers will be accumulated to the current layer. Thirdly, as shown in Figure \ref{figure:8(b)}, the MRE of $C_{l}$ layer in CNN is much larger than that of $RIC-C_{l}$ layer in RIC-CNN, because the traditional convolution is not invariant to image rotation, which leads to its feature map without rotation equivariance. Even at special angles (such as $90^{\circ}$ and $180^{\circ}$), where the interpolation and max pooling operations do not produce any calculation errors, we do not see a drop in MRE. Also, Figures \ref{figure:9(a)} and \ref{figure:9(b)} visualize some feature maps of different layers in RIC-CNN and CNN, respectively. These visualization results more intuitively confirm our above analysis. 

Finally, Table \ref{table:2} compares the efficiency of RIC-C, CNN, DEF-CNN and E(2)-CNN, the evaluation criteria include the amount of parameters and the number of images processed per second (FPS). All of these models are implemented with Pytorch. We can find that RIC-CNN and CNN contain the same number of parameters, which again shows that no additional parameters are used to achieve RIC-CNN's invariance to continous rotations. In contrast, DEF-CNN requires nearly 18\% extra trainable parameters to learn the offset field for each DEF-C layer. Meanwhile, although both RIC-C and DEF-C are implemented by using the Pytorch function \textit{DEFORM\_CONV2D}, the FPS of RIC-CNN is 1.3 times that of DEF-CNN, because the offset field of each RIC-C layer is calculated based on our rotation-invariant coordinate system in advance, and does not need to be changed when training and test the model. Indeed, the FPS of RIC-CNN is only 56\% of that of traditional CNN. However, given its good performance, this level of decline is acceptable in practice. More importantly, it can be seen that the efficiency of E(2)-CNN, which achieves the suboptimal result on the rotated test set of MNIST, is much lower than that of RIC-CNN. 
  
\begin{table}
	\caption{\label{table:2}The efficiency of RIC-C, CNN, DEF-CNN and E(2)-CNN.}
	\centering
	\begin{tabular}{p{2.8cm}p{2.5cm}p{2.5cm}p{1.8cm}}
		\toprule[1.3pt]
		Methods & Input Size & Parameters(K) & FPS(K)\\
		\toprule[1.1pt]
		E(2)-CNN\cite{47} & 29$\times$29 & 2068.3 & 0.5 \\
		CNN & 32$\times$32 & 288.6 & 77.0 \\
		DEF-CNN\cite{12} & 32$\times$32 & 340.7 & 33.3 \\
		RIC-CNN & 32$\times$32 & 288.6 & 43.5 \\
		\midrule
	\end{tabular}
\end{table}

\subsection{Real-World Image Classification}
\label{section:5.2}
\begin{figure}
	\centering
	\subfloat[Some samples from NWPU VHR-10 dataset.]
	{\includegraphics[height=35mm,width=60mm]{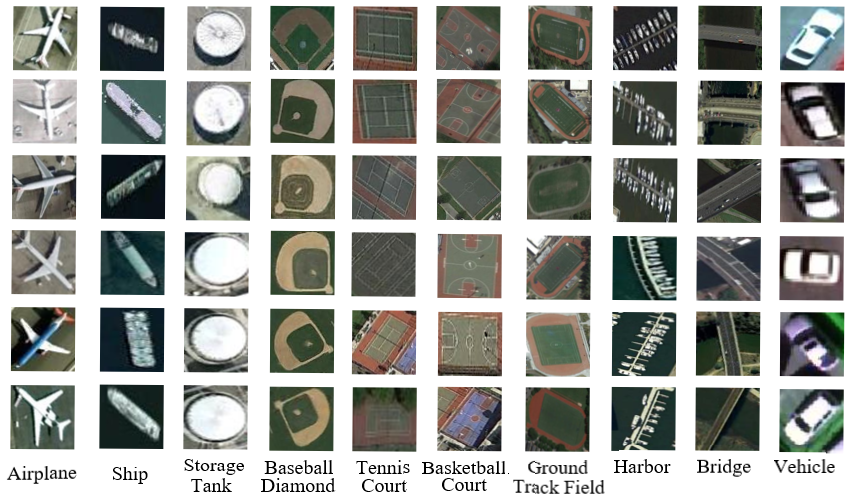}\label{figure:10(a)}\hfill}~~~~~~~
	\subfloat[Some samples from MTARSI dataset]
	{\includegraphics[height=35mm,width=60mm]{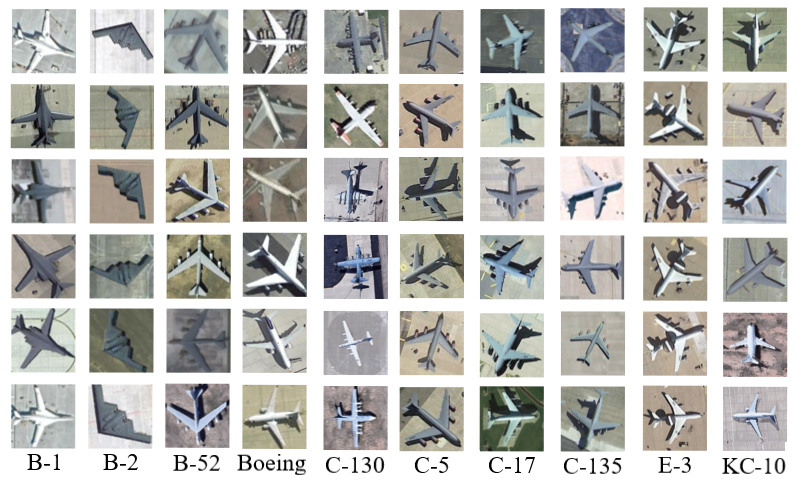}\label{figure:10(b)}\hfill}\\
	\caption{Two datasets used for real-world image classification.}\label{figure:10}
\end{figure}

In this subsection, we deploy RIC-C in three classical CNN models and conduct classification experiments on NWPU VHR-10 and MTARSI. NWPU VHR-10 is a 10-class remote sensing image dataset, and each of positive images contains at least one target to be detected \cite{17}. In previous papers, this dataset was usually used for evaluating the performance of object detection methods. In our experiment, we directly crop out all targets from positive images based on the ground truth, and use them for image classification. There are 757 airplanes, 300 ships, 655 storage tanks, 390 baseball diamonds, 524 tennis court, 159 basketball court, 163 ground track fields, 224 harbors, 124 bridges, and 598 vehicles. Each of crop regions is normalized to $64\times 64$ pixels, and some samples are shown in Figure \ref{figure:10(a)}. It is clear that targets belonging to the same class often have different orientations. This indicates that rotation-invariant features are important for classifying them. We randomly select 100 images from each class to create the training set ($10\times100$=1K), and the remaining 2894 images constitute the test set.       

Three widely used CNN architectures, VGG16\cite{14}, ResNet18\cite{15} and DenseNet40\cite{16} are selected as our baseline models. For ResNet18 and DenseNet40, the first $7\times7$ convolution layer with stride 2 is replaced with a $3\times3$ convolution layer with stride 1. This is because our input size is small. In order to test RIC-C, we replace all conventional convolutions in these models with the corresponding RIC-C, and derive RIC-VGG16, RIC-ResNet18 and RIC-DenseNet40. Again, the same protocol is used to train both CNNs and the corresponding RIC-CNNs. Specifically, the number of epochs is 100, the batch size is $10$, and models are trained with Adam optimizer while the initial learning rate is set to $10^{-4}$ ($10^{-3}$ for DenseNet40 and RIC-DenseNet40), and is multiplied by $0.5$ every 10 epochs. In the training stage, we do not use any tricks. Finally, the performance of various models is validated on the test set, and the classification results are listed in the second column of Table \ref{table:3}. 

We can see that three RIC-CNNs achieve better classification accuracies than the corresponding CNNs. For example, RIC-VGG16 outperforms VGG16 by a convincing margin of $4.42\%$. Furthermore, we reduce the size of the training set from 10$\times$100=1K to 10$\times$60=0.6K and 10$\times$30=0.3K, train six models on these two smaller sets and still test the performance of them on the original test set. The results are listed in the third and the fourth columns of Table \ref{table:3}. It can be found that as the size of the training set decreases, the accuracy gap between RIC-CNNs and the corresponding CNNs becomes larger. For example, the gap between RIC-ResNet18 and ResNet18 increases from $98.96\%-95.84\%=3.12\%$ to $98.32\%-93.08\%=5.24\%$ and $95.30\%-89.10\%=6.20\%$ when training both models using 1K, 0.6K and 0.3K images, respectively. This illustrates that, due to their natural invariance to image rotations, RIC-CNNs can learn more intrinsic information of various targets based on fewer training samples. 

\begin{table}
	\caption{\label{table:3} The classification accuracies from various models on NWPU VHR-10 dataset.}
	\centering
	\begin{tabular}{p{2.8cm}p{2.2cm}p{2.2cm}p{2.2cm}}
		\toprule[1.3pt]
		Training Data & 10$\times$100=1K & 10$\times$60=0.6K & 10$\times$30=0.3K\\
		\toprule[1.1pt]
		VGG16  & 91.64\% & 87.53\% & 82.33\%\\
		RIC-VGG16 & \textbf{96.06\%} & \textbf{93.15\%} & \textbf{91.65\%}\\
		\midrule
		ResNet18 & 95.84\% & 93.08\% & 89.10\%\\
		RIC-ResNet18 & \textbf{98.96\%} & \textbf{98.32\%} & \textbf{95.30\%}\\
		\midrule
		DenseNet40 & 97.28\% & 96.27\% & 94.30\% \\
		RIC-DenseNet40 & \textbf{99.32\%} & \textbf{98.89\%} & \textbf{97.31\%}\\
		\midrule
	\end{tabular}
\end{table}
 
In a similar fashion, we conduct the classification experiment again on MTARSI dataset. It is a new dataset for aircraft type recognition from remote sensing images, which contains 9385 images of 20 aircraft types \cite{18}. As shown in Figure \ref{figure:10(b)}, aircraft images from the same type have complicated variations in backgrounds and orientations. We also create three training sets with different size (4K, 3K and 2K) by randomly selecting 200, 150 and 100 images from each type, respectively. After training six CNNs and RIC-CNNs on various training sets, their classification accuracies on the test set (containing 9385-20$\times$200=5385 images) are listed in Table \ref{table:4}. Compared with NWPU VHR-10, MTARSI dataset contains more image categories and has stronger inter-class similarity. Thus, we can see that the performance of all six models clearly declines. However, RIC-CNNs are still significantly outperform the corresponding CNNs. For example, as the number of training images reduces from 4K to 3K and 2K, the accuracy gap between RIC-DenseNet40 and DenseNet40 increases from $95.89\%-90.50\%=5.39\%$ to $94.48\%-86.56\%=7.92\%$ and $90.04\%-79.13\%=10.91\%$. This is consistent with the experimental results on NWPU VHR-10 dataset.  

\begin{table}
	\caption{\label{table:4} The classification accuracies from various models on MTARSI dataset.}
	\centering
	\begin{tabular}{p{2.8cm}p{2.2cm}p{2.2cm}p{2.2cm}}
		\toprule[1.3pt]
		Training Data & 20$\times$200=4K & 20$\times$150=3K & 20$\times$100=2K\\
		\toprule[1.1pt]
		VGG16  & 72.10 & 71.39 & 60.15\\
		RIC-VGG16 & \textbf{83.59\%} & \textbf{82.27\%} & \textbf{72.21\%}\\
		\midrule
		ResNet18 & 85.42\% & 79.98\% & 69.84\%\\
		RIC-ResNet18 & \textbf{92.37\%} & \textbf{90.39\%} & \textbf{87.12\%}\\
		\midrule
		DenseNet40 & 90.50\% & 86.56\% & 79.13\% \\
		RIC-DenseNet40 & \textbf{95.89\%} & \textbf{94.48\%} & \textbf{90.04\%}\\
		\midrule
	\end{tabular}
\end{table}

\subsection{Image Patch Verification}
\label{section:5.3}
\begin{figure}
	\centering
	\subfloat[Some samples of positive pairs. ]
	{\includegraphics[height=30mm,width=45mm]{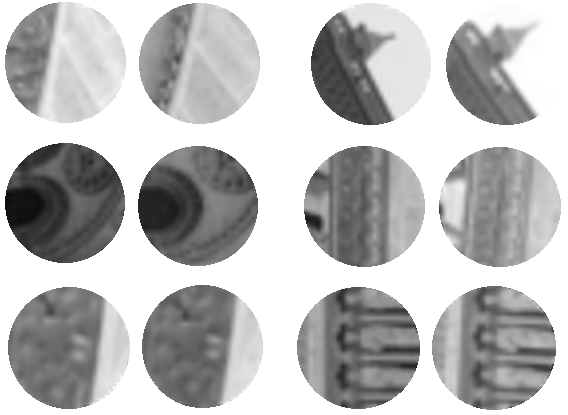}\label{figure:11(a)}\hfill}~~~~~~~~
	\subfloat[Some samples of negative pairs. ]
	{\includegraphics[height=30mm,width=45mm]{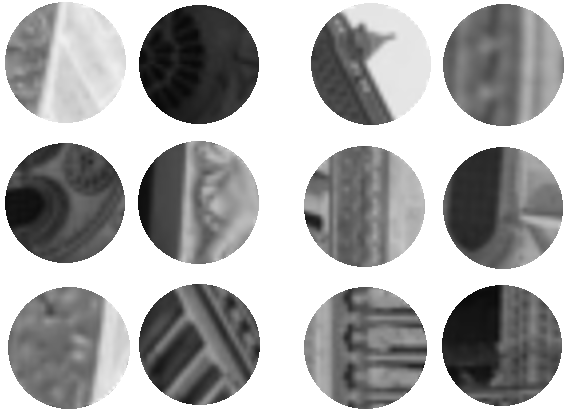}\label{figure:11(b)}\hfill}\\
	\caption{Image patch pairs from UBC benchmark dataset.}\label{figure:11}
\end{figure}

In our last experiment, UBC benchmark dataset is used to evaluate RIC-CNN's performance in patch verification task. This dataset consists of three subsets, Liberty, Yosemite and Notredame, and in each of them, there are more than 500k patch pairs, including positive and negative pairs \cite{20}. Two patches in a positive pair correspond to the same 3D keypoint, while two patches in a negative pair correspond to different 3D keypoints. Some patch pairs from UBC benchmark dataset are shown in Figure \ref{figure:11}. Each patch is scaled to $32\times32$ pixels, and we just use the content located in the area of the inscribed circle. In patch verification, we need to extract the features of two patches in a pair, and then judge whether they constitute a positive pair based on the Euclidean distance between their features. 
 
A simple CNN architecture is designed to extract image patch features. It is made of 4 convolutional layers with 32, 64, 128 and 256 kernels of size $3\times3$, respectively; one $2\times2$ max pooling after each of first three layers; one $4\times4$ max pooling after the last convolution layer. All activation functions are Tanh. The final output is a 256-dimensional feature vector. Again, we replace all $3\times3$ convolutional operations with $3\times3$ RIC-C and obtain the corresponding RIC-CNN. Both CNN and RIC-CNN are trained with triple loss, and the training set contains 1M triplets from one subset. To form a triplet for training, a positive patch pair and a single patch from another 3D keypoint are randomly chosen. In fact, we basically follow the training protocol proposed in \cite{19}. For the optimization, the Stochastic Gradient Descend is used, and the initial learning rate and weight decay are $0.1$ and $10^{-4}$, respectively. The models are trianed with 100 epochs and the learning rate is multiplied by $0.8$ every 10 epochs. 

We train two models on one subset, and validate their performance on the other two subsets. A test set includes 100K patch pairs chosen from one subset (50K positive pairs and 50K negative pairs). For each of patch pairs, the distance between two patch features is calculated, and then all patch pairs are sorted based on feature distances (ascending order). Clearly, patch pairs with small feature distances are regarded as positive pairs. We use the false positive rate at 95\% true positive rate (FPR95) as the metric to evaluate the performance of various models. In fact, based on gradient information, each patch in UBC benchmark dataset has been normalized to a standardized orientation. The authors also required that the orientation difference of two patches in a positive pair is less than $22.5^{\circ}$. However, as stated in Section \ref{section:3}, the dominant orientation assignment is an error-prone process. Thus, in practice, there are more severe rotation variation between image patches. To simulate this and increase the difficulty of our task, two patches in a pair are rotated by different random angles $\theta$. By limiting the variation range of $\theta$ $\left([-30^{\circ},30^{\circ}],[-60^{\circ},60^{\circ}],[-90^{\circ},90^{\circ}],[-120^{\circ},120^{\circ}],[-150^{\circ},150^{\circ}],[-180^{\circ},180^{\circ}]\right)$, for each original test set, we further obtain six additional test sets. These rotated test sets are more challenging than the original one.   

\begin{figure}
	\centering
	\subfloat[Notredame-Liberty]
	{\includegraphics[height=50mm,width=60mm]{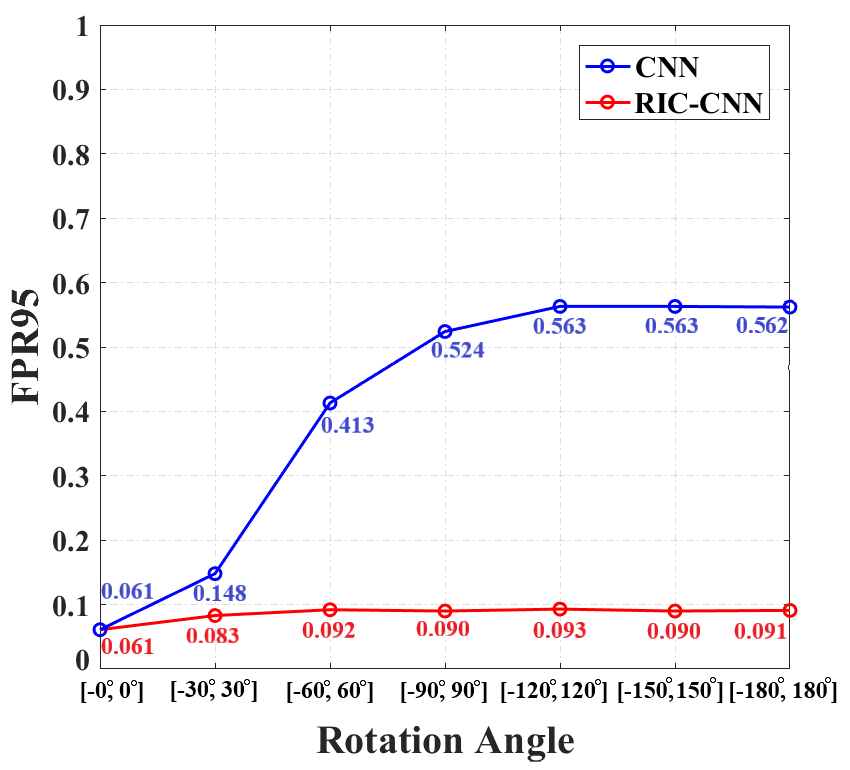}\label{figure:12(a)}\hfill}~~~~~
	\subfloat[Notredame-Yosemite]
	{\includegraphics[height=50mm,width=60mm]{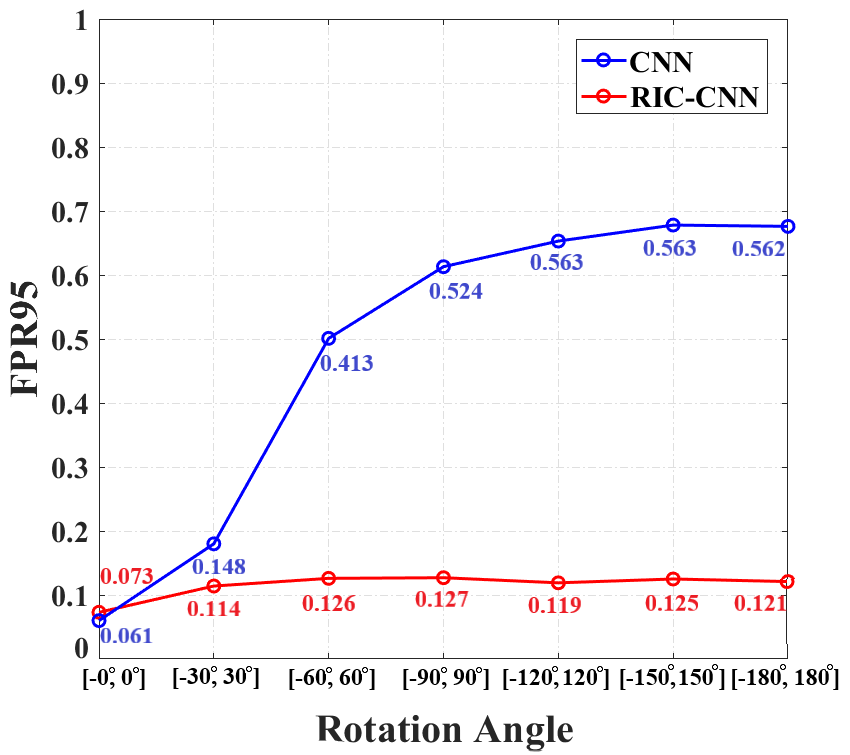}\label{figure:12(b)}\hfill}\\
	\subfloat[Yosemite-Liberty]
	{\includegraphics[height=50mm,width=60mm]{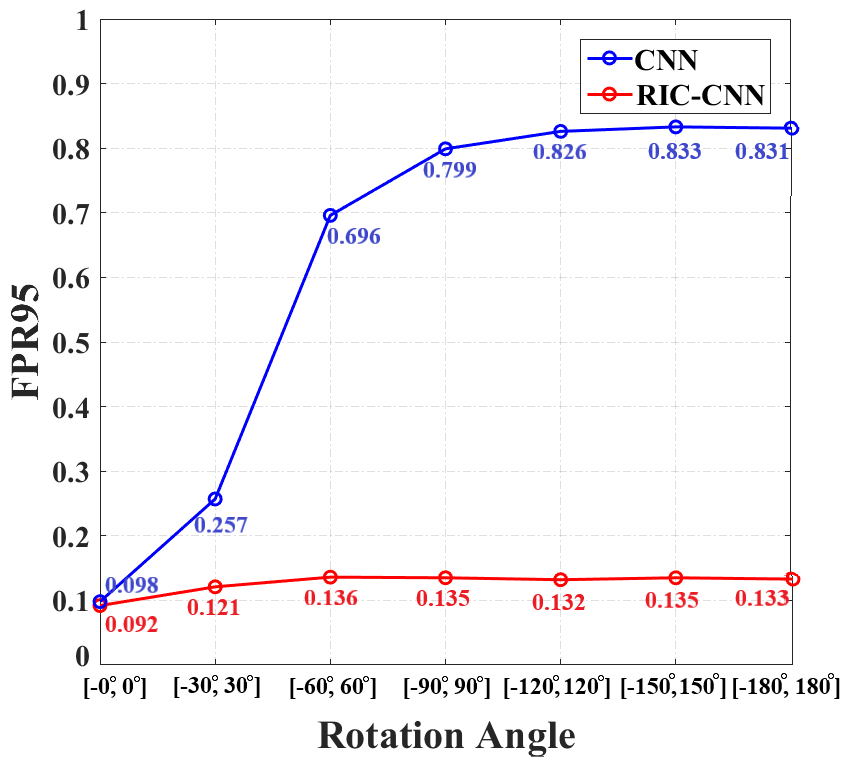}\label{figure:12(c)}\hfill}~~~~~
	\subfloat[Yosemite-Notredame]
	{\includegraphics[height=50mm,width=60mm]{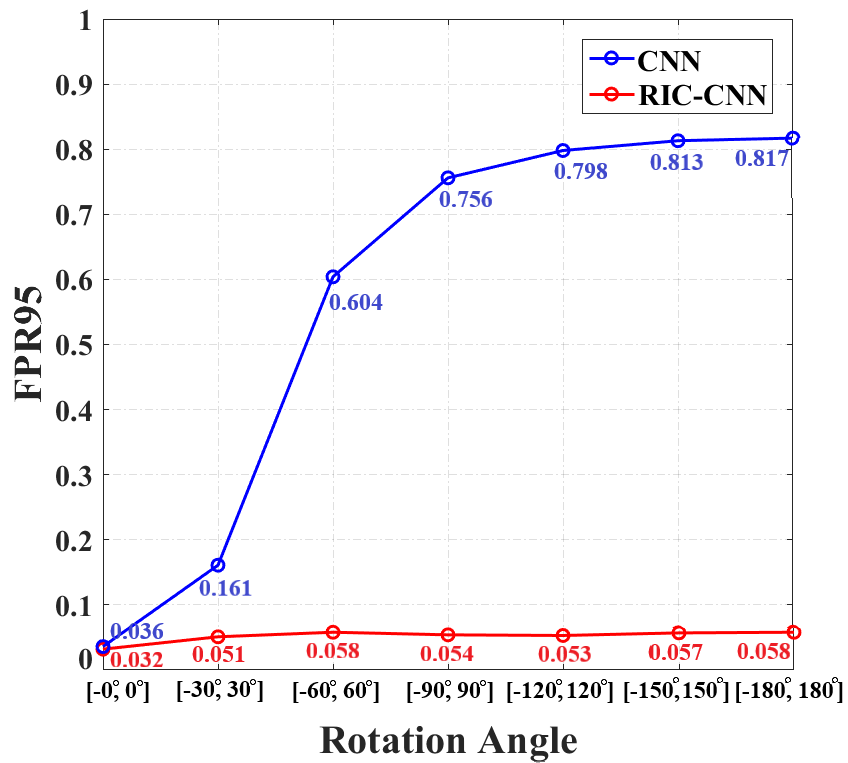}\label{figure:12(d)}\hfill}
	\caption{The results for four combinations of training and testing using Liberty, Yosemite and Notredame subsets. Note that Notredame-Liberty means that we train the models on Notredame and test their performance on Liberty.}\label{figure:12}
\end{figure}

The results for four combinations of training and testing using three subsets are shown in Figure \ref{figure:12}. Using original test sets (without applying additional rotations), we can find that RIC-CNN slightly outperforms CNN on three cases (except for Notredame-Yosemite, see Figure \ref{figure:12(b)}). However, CNN is extremely sensitive with respect to image rotations. Taking the combination Yosemite-Notredame as an example (see Figure \ref{figure:12(d)}), when test patches are rotated by $\theta\in[-60^{\circ},60^{\circ}]$, the FPR95 from using CNN increases from $0.036$ to $0.604$. This means that the model is almost out of order, because more than half of negative pairs are misclassified. In contrast, RIC-CNN are much more robust. Even for heaviest rotation variation $\theta\in[-180^{\circ},180^{\circ}]$, it still achieves $0.058$ FPR95 (just increases 0.026). This again shows that, without data augmentation and extra training parameters, RIC-C greatly enhances the rotation invariance of CNN models while maintaining CNNs original performance.  

\section{Conclusions and Future Work}
\label{section:6}
A new convolutional operation RIC based on rotation-invariant coordinate system is proposed in this paper, which is naturally invariant to continous rotations around the input center. We analyze the connection between RIC-C and DEF-C, and find a simple but efficient method to implement RIC-C using Pytorch. Given a standard CNN, we can obtain a RIC-CNN by just replacing all standard convolutions with the corresponding RIC-C. Based on MNIST dataset, we first verify the rotation invariance of RIC-CNN. When training RIC-CNN and other existing rotation-invariant CNN models on the original training set without data augmentation, RIC-CNN achieves the best classification accuracy on the rotated test set of MNIST. Furthermore, we deploy RIC-C to several classical CNN architectures, and conduct image classification and patch verification on several real image datasets. The experimental results show that RIC-C can be easily used as drop in replacement for standard convolutions, and obviously enhances the rotation invariance of CNN models designed for various applications. In the future, we plan to apply RIC-CNN in more practical tasks, such as rotation-invariant face detection, plankton classification and so on. In addition, RIC-C can only handle arbitrary rotations around the input center. We will design other solutions to break this limitation.

\begin{ack}
This work has partly been funded by the Academy Professor Project of the Academy of Finland (EmotionAI, Grant No.336116), the National Key R\&D Program of China (No.2017YFB1002703) and the National Natural Science Foundation of China (Grant No.60873164, 61227802 and 61379082)
\end{ack}

\end{document}